\documentclass[lettersize,journal]{IEEEtran}
\usepackage{amsmath,amsfonts}
\usepackage{algorithmic}
\usepackage{algorithm}
\usepackage{setspace}
\usepackage{array}
\usepackage{textcomp}
\usepackage{stfloats}
\usepackage{url}
\usepackage{verbatim}
\usepackage{graphicx}
\usepackage{subcaption}
\usepackage{siunitx}
\usepackage{subcaption}
\usepackage{cite}
\usepackage{booktabs}
\usepackage{multirow}
\usepackage{hyperref}
\hypersetup{hidelinks,
	colorlinks=true,
	allcolors=black,
	pdfstartview=Fit,
	breaklinks=true}

\def\BibTeX{{\rm B\kern-.05em{\sc i\kern-.025em b}\kern-.08em
    T\kern-.1667em\lower.7ex\hbox{E}\kern-.125emX}}
\usepackage{balance}

\usepackage{threeparttable}

\usepackage{makecell}

\usepackage{tabularx}

\begin{document}
\title{VAN-AD: Visual Masked Autoencoder with Normalizing Flow for Time Series Anomaly Detection}
\author{PengYu~Chen,~Shang~Wang,~Xiaohou~Shi,~Yuan~Chang,~Yan~Sun,~and~Sajal~K. Das,~\IEEEmembership{Fellow,~IEEE}
\thanks{
This work was supported by the National Natural Science Foundation of China under Grant 62272052. Yan Sun is the corresponding author.}
\thanks{
Pengyu Chen, Shang Wang and Yan Sun are with the School of Computer Science (National Pilot Software Engineering School), Beijing University of Posts and Telecommunications, Beijing 100876, China (e-mail: penychen@bupt.edu.cn; wangshang@bupt.edu.cn; sunyan@bupt.edu.cn).

Xiaohou Shi and Yuan Chang are engineers of China Telecom Research Institute Beijing, China (e-mail: shixh6@chinatelecom.cn;
changy8@chinatelecom.cn).

Sajal K. Das is with the Department of Computer Science, Missouri University
of Science and Technology, Rolla, MO 65409 USA (e-mail: sdas@mst.edu).}
}

\markboth{Journal of \LaTeX\ Class Files,~Vol.~18, No.~9, September~2020}%
{How to Use the IEEEtran \LaTeX \ Templates}

\maketitle
\begin{abstract}
Time series anomaly detection (TSAD) is essential for reliable and secure IoT-enabled service systems. 
However, dataset-specific training limits the generalization of existing methods, especially when target-domain training data are scarce. 
Although foundation models offer a promising direction, current attempts based on large language models or large-scale time-series pre-training still suffer from cross-modal gaps or in-domain heterogeneity.
In this paper, we investigate the applicability of large-scale vision models to TSAD. Specifically, we adapt a visual Masked Autoencoder (MAE) pre-trained on ImageNet to the TSAD task. However, directly transferring MAE to TSAD introduces two key challenges: over-generalization and limited local perception. To address these challenges, we propose VAN-AD, a framework that adapts an ImageNet-pre-trained visual MAE to TSAD, which combines adaptive reconstruction post-processing and flow-based density estimation. To alleviate the over-generalization issue, we design an Adaptive Distribution Modulation Module (ADMM) as a training-free post-processing operation, which recalibrates the MAE reconstruction using the local statistics of the input window to amplify discrepancies caused by abnormal patterns. To overcome the limitation of local perception, we further develop a Normalizing Flow Module (NFM), which combines MAE with normalizing flow to estimate the probability density of the current window under the global distribution. Extensive experiments on nine real-world datasets demonstrate that VAN-AD consistently outperforms existing state-of-the-art methods across multiple evaluation metrics. We make our code and datasets available at \href{https://github.com/PenyChen/VAN-AD}{https://github.com/PenyChen/VAN-AD}.

\end{abstract}


% This figure illustrates over-generalization and local-perception.
\begin{figure*}[t]
    \centering
    \begin{subfigure}{0.48\linewidth}
        \centering
        \includegraphics[width=\linewidth]{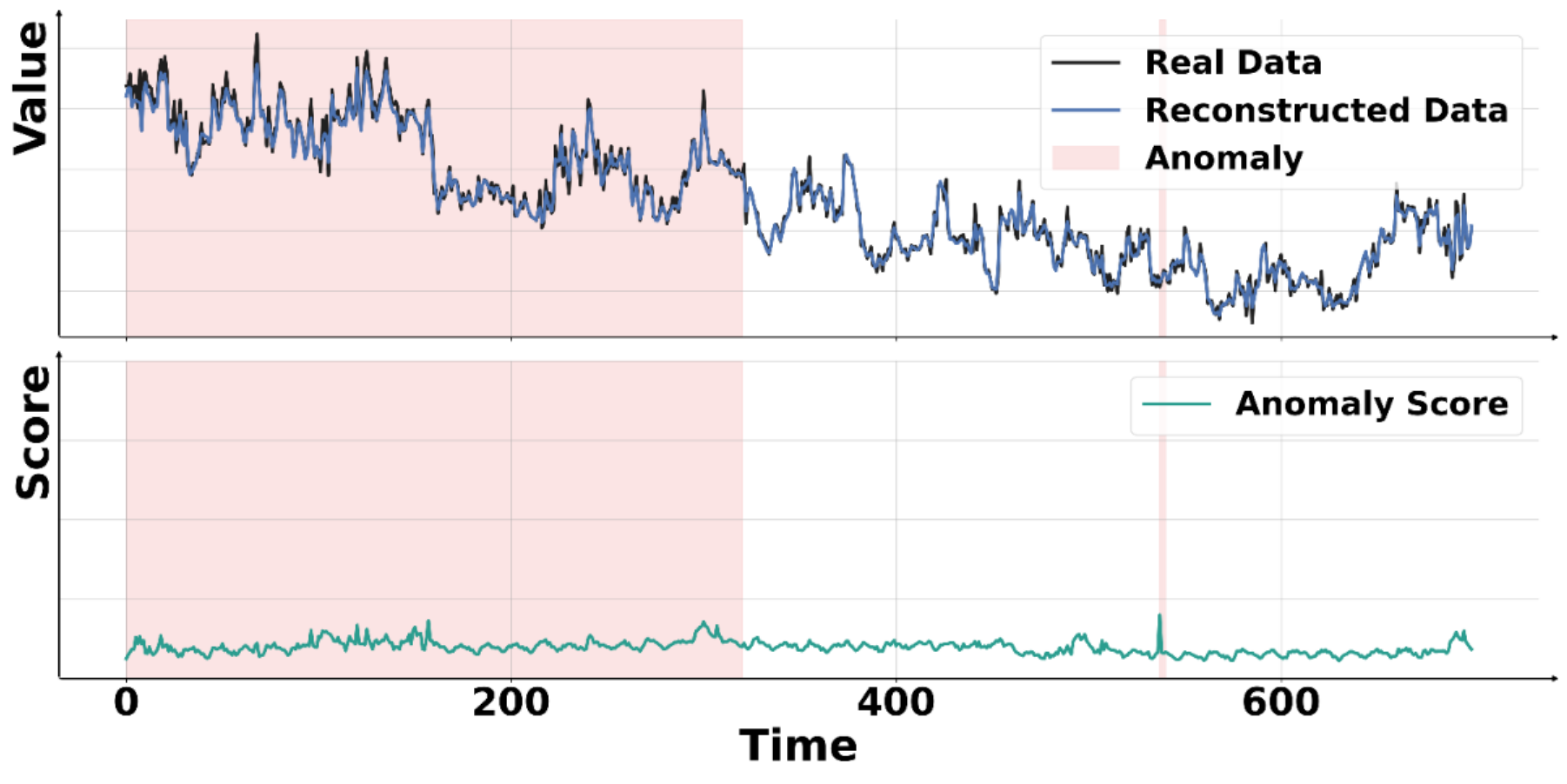}
        \caption{over-generalization}
        \label{fig:over_generalization}
    \end{subfigure}
    \hfill
    \begin{subfigure}{0.48\linewidth}
        \centering
        \includegraphics[width=\linewidth]{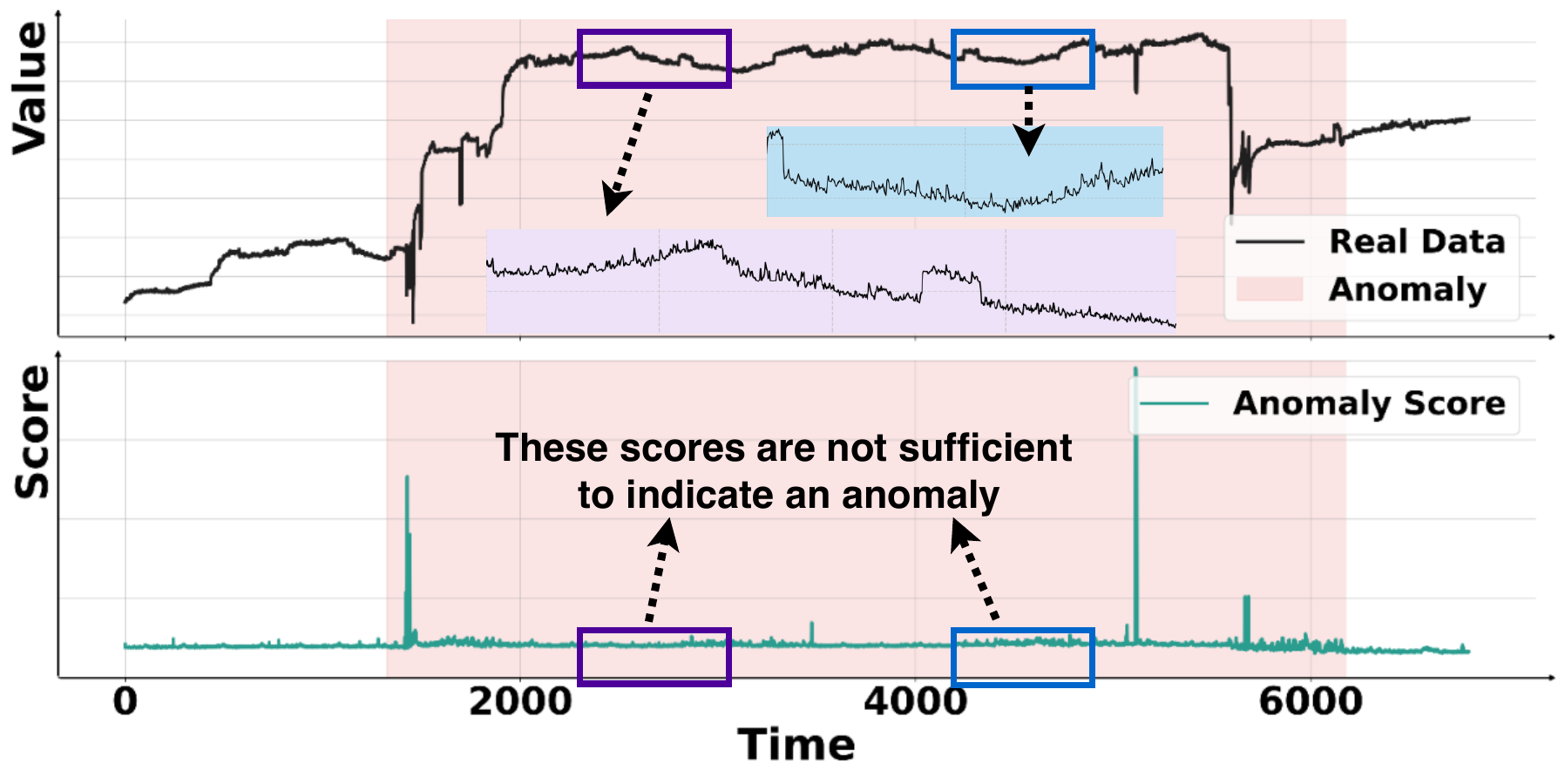}
        \caption{local-perception}
        \label{fig:local_perception}
    \end{subfigure}
    \caption{Examples of the over-generalization and local-perception issues of MAE in TSAD on the PSM dataset. The red background indicates anomalous segments. Fig. \ref{fig:over_generalization} shows that both normal and anomalous sequences can be well reconstructed, resulting in similarly low anomaly scores. Fig. \ref{fig:local_perception} illustrates that when a window is fully occupied by anomalies, variations within the window fail to reveal abnormality, leading to low anomaly scores.
}
    \label{fig:mae_issues}
\end{figure*}

\section{Introduction}
\IEEEPARstart{I}{n} IoT-enabled service systems, time series anomaly detection (TSAD) is critical for maintaining system reliability and security, as it helps identify device failures, sensor malfunctions and malicious operations from temporal data~\cite{zamanzadeh2024deep}. With the rapid proliferation of IoT devices and advances in data acquisition technologies, massive volumes of time series data are being continuously generated, posing increasing demands on effective TSAD. Consequently, deep learning methods have been widely adopted and have shown strong capability in modeling complex temporal patterns~\cite{wucatch,licrossad,yinscatterad}. However, these methods still rely heavily on dataset-specific training and thus generalize poorly across diverse target scenarios~\cite{shentu2025towards}. This limited transferability hinders their practical deployment in IoT service systems, where data are often scarce and large-scale data sharing is restricted by annotation costs and privacy concerns.

Against this background, recent studies have increasingly turned to foundation models for TSAD. Existing efforts mainly follow two paradigms: TS-based and Text-based pre-training methods. TS-based methods build foundation models on large-scale time series corpora from diverse domains or synthetic generation. For example, DADA~\cite{shentu2025towards} improves generalization by synthesizing anomalous samples during pre-training. However, the scarcity of real-world anomalous samples makes it difficult to construct high-quality large-scale pre-training datasets, thereby limiting the scalability of TS-based methods.
In contrast, Text-based methods attempt to transfer the knowledge of large language models (LLMs) pre-trained on massive text corpora to TSAD. For instance, AnomalyLLM~\cite{liu2024large} transfers the knowledge of GPT-2 to TSAD via knowledge distillation. Nevertheless, the substantial modality gap between textual and temporal data raises concerns about whether such methods can effectively capture the underlying temporal dynamics of time series~\cite{zhoucan,dong2024can}. These limitations motivate the exploration of foundation models that are more structurally aligned with temporal data.

In light of these limitations, pre-trained vision models have recently shown considerable promise for time series analysis. Compared with discrete textual data, images and time series are both continuous and share common structural characteristics, such as long-range trends, periodicity, and abrupt state changes~\cite{chen2025visionts,shen2025visionts++,li2023time}. This structural alignment suggests that pre-trained vision models can offer a more suitable source for transfer learning in time series tasks than Text-based methods.
For example, VisionTS~\cite{chen2025visionts} and VisionTS++~\cite{shen2025visionts++} convert time series segments into two-dimensional grayscale images and leverage representations learned by the ImageNet-pre-trained visual masked autoencoder for zero-shot forecasting, while ViTST~\cite{li2023time} demonstrates the effectiveness of Vision Transformers for time series classification through line-chart rendering. However, existing studies on vision-based transfer for time series have primarily focused on forecasting and classification, leaving a clear research gap in adapting pre-trained vision models to TSAD.
Motivated by these advances, we investigate the applicability of the ImageNet-pre-trained visual masked autoencoder~\cite{he2022masked} to TSAD, which we refer to as MAE for simplicity in the following. Pre-trained on large-scale natural image datasets, MAE shows strong potential as a transferable backbone for TSAD, and we leverage its reconstruction behavior to identify abnormal temporal observations.

Nevertheless, directly transferring a pre-trained MAE to TSAD remains nontrivial. A major challenge stems from its strong reconstruction capability, which can lead to over-generalization: anomalous observations may be reconstructed nearly as well as normal ones, thereby reducing the distinction between normal and abnormal patterns, as illustrated in Fig.~\ref{fig:over_generalization}. 
Existing studies often address this issue by introducing additional trainable modules (e.g., memory banks) to improve anomaly discrimination~\cite{shen2025learn,song2023memto}. Although effective in some cases, such solutions typically require target-domain retraining and extra parameter optimization, thereby weakening the direct transferability of MAE and limiting the benefit of its rich visual priors. Therefore, we aim to amplify the reconstruction discrepancy between normal and anomalous samples by post-processing the outputs of the MAE backbone, rather than modifying the backbone or introducing additional trainable parameters.

A further challenge lies in the widely adopted fixed sliding window scheme in TSAD, which gives rise to a local perception issue: the model can only capture patterns within a limited temporal window, while lacking awareness of broader temporal context. As shown in Fig.~\ref{fig:local_perception}, when an anomaly occupies an entire input window, the model may still reconstruct it well and treat it as a normal temporal variation. This issue is particularly problematic in IoT security scenarios, where sustained attacks or persistent device malfunctions may span an entire window and thus be mistaken for normal operating patterns. Existing studies typically address this issue through two main strategies: enlarging the receptive field ~\cite{haq2025transnas} and introducing explicit context modeling mechanisms ~\cite{park2026paano}. However, the former substantially increases computational overhead, especially for Transformer-based ~\cite{vaswani2017attention} backbones, whereas the latter often requires additional context-encoding modules and thus increases inference complexity. Therefore, instead of introducing another heavy-weight temporal encoder, we aim to design a lightweight mechanism that provides the model with explicit information about the global distribution of the input time series.

To address these challenges, we propose \textbf{VAN-AD}: a novel \textbf{V}isual Masked \textbf{A}utoencoder with \textbf{N}ormalizing Flow for Time Series \textbf{A}nomaly \textbf{D}etection. We first encode multivariate time series into variable-time heatmaps and use checkerboard mask to construct complementary image pairs for MAE reconstruction. To explicitly expose anomalous deviations and mitigate over-generalization, we design an Adaptive Distribution Modulation Module (ADMM) as a training-free post-processing operation, which recalibrates the MAE reconstruction using the local statistics of the input window to enhance reconstruction discrepancies between normal and anomalous samples. We further introduce a Normalizing Flow Module (NFM) to model the probability distribution of normal patterns from MAE output features, thereby providing explicit global distributional information beyond the fixed sliding window scheme. By combining local reconstruction information from MAE with global distributional modeling from NFM, VAN-AD achieves more accurate anomaly detection performance.

Our main contributions are summarized as follows:
\begin{itemize}
    \item We propose VAN-AD, a novel framework that adapts an ImageNet-pre-trained visual MAE to TSAD. Experiments on nine real-world datasets show that VAN-AD consistently outperforms existing baselines. In particular, it achieves notable gains over the representative TS-based model DADA, with average improvements of 21.1\% in VUS-ROC and 25.1\% in VUS-PR.
    
    \item We design an Adaptive Distribution Modulation Module (ADMM) as a training-free post-processing operation to recalibrate MAE reconstructions using local input statistics, thereby mitigating the over-generalization issue.
    
    \item We introduce a Normalizing Flow Module (NFM) to model the global probability distribution of normal patterns, thereby compensating for the local perception issue.
\end{itemize}

\section{Related Work}
This section summarizes related work on time series anomaly detection, vision-based time series analysis, and MAE-based methods.
\subsection{Time Series Anomaly Detection}
Traditional TSAD methods can be classified into non-learning\cite{breunig2000lof,he2003discovering} and machine learning\cite{ramaswamy2000efficient,shyu2003novel}. In recent years, deep learning methods have achieved remarkable success in TSAD. These methods can be broadly classified into three main categories: forecasting-based\cite{deng2021graph}, reconstruction-based\cite{wucatch,xie2025multivariate}, and contrastive-based approaches\cite{yang2023dcdetector,kim2025causality}.
Forecasting-based methods identify anomalies through prediction errors. For example, GDN\cite{deng2021graph}, captures inter-variable dependencies and uses deviations between predicted and observed values as anomaly signals. Reconstruction-based methods detect anomalies according to reconstruction error, under the assumption that normal patterns are easier to reconstruct than abnormal ones. Representative methods such as CATCH\cite{wucatch} and MTSCID\cite{xie2025multivariate} improve detection by jointly modeling temporal and frequency information. 
Contrastive-based methods focus on learning discriminative representations, where normal patterns are encouraged to cluster while anomalous patterns are separated. For instance, CAROTS\cite{kim2025causality} incorporates causal information into contrastive learning to enhance anomaly discrimination.

Although these methods achieve strong performance on specific datasets, they generally require dataset-specific training and therefore are difficult to transfer to new scenarios where data are scarce or privacy constraints are strict. As a result, foundation models for TSAD have recently attracted increasing attention. According to the modality of pre-training data, existing approaches can be broadly divided into TS-based and Text-based methods.
Among TS-based based methods, DADA\cite{shentu2025towards} constructs anomalous samples during pre-training through heuristic anomaly injection and explicitly improves the discriminability between normal and abnormal patterns via Adaptive Bottlenecks and Dual Adversarial Decoders. Other foundation models, such as Time-MoE\cite{xiaoming2025time}, Chronos\cite{ansari2024chronos}, and Timer\cite{liu2024timer}, are pre-trained on diverse real-world time series datasets collected from multiple domains, and assess the anomaly level of each time point based on prediction errors.

Despite these advances, high-quality anomaly detection pre-training datasets remain scarce, as anomalous patterns are inherently rare. 
This limits the ability of existing TS-based methods to handle complex industrial evolution patterns.
Text-based methods provide an alternative direction. GPT4TS\cite{zhou2023one} adapts LLMs to TSAD through fine-tuning. AnomalyLLM\cite{liu2024large}, in contrast, uses GPT-2 as a teacher model to supervise a student network through knowledge distillation, and identifies anomalies according to the discrepancy between the teacher and student outputs. Zhang et al.\cite{zhang2024large} further explores the use of GPT-4 and Claude-2 for anomaly detection and requires the models to provide explanations for their decisions. Nevertheless, because LLMs are inherently developed for discrete textual data, whereas time series are continuous signals, the substantial modality gap makes it difficult for such methods to capture the underlying evolution patterns of time series. Consequently, the effectiveness of Text-based methods for TSAD has been increasingly questioned\cite{zhoucan,dong2024can}.

\subsection{Vision-based Time Series Analysis}
In recent years, Vision-based methods have shown considerable potential in time series analysis. AST\cite{gong2021ast} employs DeiT\cite{touvron2021training} for time series classification, while ViTST\cite{li2023time} further explores this direction by leveraging pre-trained Vision Transformers. VisionTS\cite{chen2025visionts} demonstrates that the visual Masked Autoencoder exhibits remarkable zero-shot transferability in time series forecasting, and its follow-up work, VisionTS++\cite{shen2025visionts++}, further enhances the representational capacity of cross-modal foundation models through continued pre-training of the visual backbone. In addition, Ni et al.\cite{ni2025harnessing} systematically review the potential of vision models for time series analysis, highlighting that two-dimensional image representations can more naturally preserve the structural information of complex fluctuations. Zhao et al.\cite{zhao2025images} further provide a comprehensive evaluation of large-scale vision models for time series processing, confirming the strong inductive bias introduced by visual pre-training.

Despite these encouraging results in time series forecasting and classification, the application of vision models to TSAD remains at an early stage. Existing studies have not yet provided effective designs for adapting vision foundation models to anomaly detection, leaving substantial room for investigating an anomaly detection framework based on vision models. In this work, we explore this direction by applying a pre-trained MAE to TSAD.

\subsection{Masked Autoencoder-based Methods for TSAD}
Although directly adapting ImageNet-pre-trained visual MAEs to TSAD remains underexplored, the idea of masked reconstruction has been adopted in several time-series-native anomaly detection methods. 
MAD~\cite{fu2022mad} introduces masked reconstruction as a self-supervised anomaly detection task for multivariate time series, where partially masked observations are reconstructed to learn normal temporal patterns. 
TFMAE~\cite{fang2024temporal} designs temporal-frequency masked autoencoders, using temporal and frequency masking strategies to reduce abnormal bias and adopting a contrastive criterion to alleviate distribution shifts. 
MMA~\cite{tang2024mlp} combines the masked autoencoder paradigm with an MLP-Mixer backbone to handle larger input windows for detecting long-duration subsequence anomalies, and further incorporates contrastive learning and dynamic anomaly filtering. 
TSMAE~\cite{liu2025time} applies masked autoencoding to process anomaly detection by randomly masking input variables and reconstructing them with an encoder-decoder structure, aiming to enhance variable-correlation modeling. 

Different from these time-series-native MAE-based methods, VAN-AD focuses on adapting a large-scale vision model to TSAD. 
It explores the potential of an ImageNet-pre-trained visual MAE as a frozen reconstruction backbone for time series anomaly detection. 
Unless otherwise specified, MAE refers to the ImageNet-pre-trained visual MAE in the following sections.

\section{Preliminaries}
This section first introduces the concept of time series anomaly detection, and then summarizes the definition of visual masked autoencoders and normalizing flow.
\subsection{Time Series Anomaly Detection}
Given a multivariate time series $X=\left(x_1, x_2, \ldots, x_T\right)\in\mathbb{R}^{T\times C}$, where $T$ is the sequence length and $C$ is the number of variables, time series anomaly detection aims to predict an anomaly label sequence $Y=\left(y_1, y_2, \ldots, y_T\right)\in\{0,1\}^{T}$, where $y_t$ indicates whether the observation $x_t$ at time step $t$ is anomalous.

% Architecture of VAN-AD
\begin{figure*}[t]
    \centering
    \includegraphics[width=1.0\linewidth]{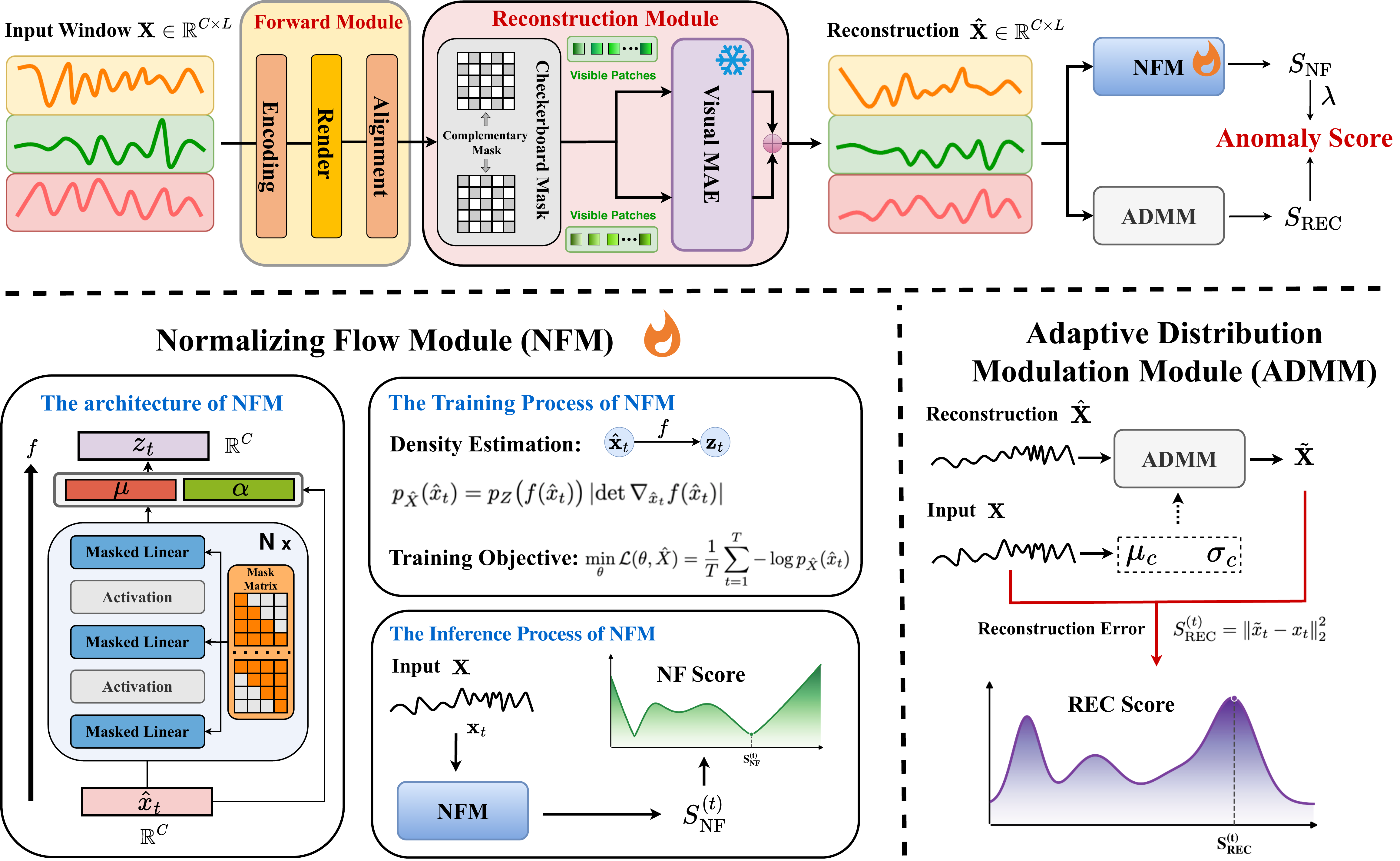}
    \caption{The overall architecture of VAN-AD. (1) Forward Module: transforms the input time series into a format compatible with MAE. (2) Reconstruction Module: generates a pair of complementary masked images using the checkerboard mask strategy, and reconstructs the masked regions through MAE to obtain the reconstructed time series. (3) ADMM: post-processes the MAE reconstruction using local input statistics to amplify reconstruction discrepancies. (4) NFM: applies normalizing flow to the reconstructed sequence to model the probability density of the current window.}
    \label{fig:van-ad}
\end{figure*}

\subsection{ImageNet-pre-trained Visual Masked Autoencoders}
MAE\cite{he2022masked} is a computer vision foundation model pre-trained on ImageNet\cite{deng2009imagenet} in a self-supervised manner. Its core idea is to learn image representations through a mask and reconstruct pretext task. Specifically, the input image $I$ is first divided into a sequence of non-overlapping patches, after which a very high masking ratio (typically up to 75\%) is applied so that only a small subset of visible patches is retained as input to the encoder. This asymmetric design is central to MAE: the encoder processes only the visible patch sequence and maps it into a latent feature space, while the decoder takes the encoded latent representations together with mask tokens indicating the missing positions and attempts to reconstruct the pixel values of the original image. By minimizing the mean squared error (MSE) between the reconstructed output and the original image over the masked regions during training, MAE is encouraged not only to capture local details but also to infer missing information from the global contextual structure.

As a reconstruction vision model, MAE can be naturally adapted to TSAD. Through pre-training on large-scale natural image datasets, it learns rich continuous structural patterns, many of which are also common in time series, such as trends and seasonality. This structural similarity suggests that a pre-trained MAE may provide transferable priors for modeling normal temporal dynamics, and thus reconstruct normal sequences more effectively than anomalous ones.

\subsection{Normalizing Flow}
Normalizing Flow is an unsupervised density estimation technique that transforms a complex data distribution into a simple latent distribution through a sequence of invertible transformations\cite{dai2022graph,zhou2024label}. Its core idea is to achieve exact likelihood modeling of the target data based on the change of variables theorem.

Let the observed variable $x \in \mathbb{R}^{d}$ follow an unknown distribution $p_X(x)$. Normalizing Flow defines a bijective function $f:\mathcal{X}\rightarrow\mathcal{Z}$, which maps $x$ to a latent variable $z \in \mathbb{R}^{d}$ following a known simple distribution, such as a standard Gaussian distribution $p_Z(z)=\mathcal{N}(0,I)$, i.e., $z=f(x)$.

According to the change of variables theorem, the probability density of the observed data $x$ can be written as
\begin{equation}
p_X(x)=p_Z\bigl(f(x)\bigr)\left|\det \nabla_x f(x)\right|
\label{eq:nf_density},
\end{equation}
where $\nabla_x f(x)$ denotes the Jacobian matrix of the transformation $f$ at $x$.

To enhance the representational capacity, the transformation $f$ is constructed as a composition of $K$ simple invertible mappings:
\[
f = f_K \circ f_{K-1} \circ \cdots \circ f_1.
\]

Let $z_0=x$ and $z_K=z$. Then, the log-likelihood of the overall transformation can be expressed as
\begin{equation}
\log p_X(x)=\log p_Z(z_K)+\sum_{k=1}^{K}\log\left|\det \nabla_{z_{k-1}} f_k(z_{k-1})\right|.
\label{eq:nf_loglikelihood}
\end{equation}

This mechanism enables the model to capture the distributional characteristics of normal observations by directly maximizing their log-likelihood.

\section{Methodology}
This section overviews the architecture of VAN-AD and then details each module in the model.

\subsection{Overview}
Fig. \ref{fig:van-ad} illustrates the overall architecture of VAN-AD. Given a multivariate time series $X \in \mathbb{R}^{C \times L}$, we first transform it into a two-dimensional image through the Forward Module and construct a pair of complementary masked images using checkerboard mask. A pre-trained visual MAE is then applied to recover the masked regions, and the reconstructed image is subsequently converted back into a multivariate time series. To enlarge the discrepancy between the original and reconstructed sequences, we introduce ADMM, which post-processes the MAE outputs using the local statistics of the input window. Meanwhile, NFM is designed to model the probability density of normal patterns, allowing the model to incorporate the density of the current window under the global distribution into anomaly detection.

\subsection{The Forward Module}

Following the preprocessing convention of pre-trained vision backbones, we first normalize the time series using the mean and standard deviation computed from the training set, and apply the same statistics to the test set to avoid data leakage, which is a standard practice in current TSAD~\cite{qiu2025tab}. Therefore, the input window used in VAN-AD is a standardized sequence \(X\in\mathbb{R}^{C\times L}\).

To preserve variable information in multivariate time series, we adopt a heatmap encoding strategy. Specifically, we directly regard \(X\) as a variable-time grayscale map, where the row index corresponds to the variable dimension and the column index corresponds to the temporal dimension. The pixel intensity at location \((c,t)\) is assigned as
$I_{\mathrm{raw}}(c,t)=X_{c,t}$, where \(I_{\mathrm{raw}}\in\mathbb{R}^{C\times L}\).

To make it compatible with the pre-trained MAE, we replicate \(I_{\mathrm{raw}}\) across all three channels to obtain a grayscale RGB image:
\begin{equation}
I_{\mathrm{gray}}(c,t,k)=I_{\mathrm{raw}}(c,t), \quad k=1,2,3,
\end{equation}
where \(I_{\mathrm{gray}}\in\mathbb{R}^{C\times L\times3}\). Since the pre-trained MAE uses a fixed input resolution of \(224\times224\), \(I_{\mathrm{gray}}\) is further resized to this resolution through bilinear interpolation, yielding the final input image \(I\in\mathbb{R}^{224\times224\times3}\).

\subsection{The Reconstruction Module}
In this paper, we adopt a checkerboard mask strategy to construct a pair of complementary masked images. Each image is reconstructed from its visible regions, allowing the model to more fully capture both temporal dependencies and inter-variable correlations.

Specifically, the input image $I$ is first divided into a set of non-overlapping patches. Let the patch size be $P \times P$. The image is then partitioned into an $N \times N$ grid, where $N = 224 / P$. Each patch is indexed by coordinates $(i,j)$, where $i,j \in \{0,1,\ldots,N-1\}$. Based on this patch grid, we define a checkerboard mask as
\begin{equation}
M(i,j)=
\begin{cases}
1, & \text{if } (i+j)\ \text{is even},\\
0, & \text{if } (i+j)\ \text{is odd},
\end{cases}
\qquad
\bar{M}(i,j)=1-M(i,j),
\label{eq:checkerboard_mask}
\end{equation}
where $M$ and $\bar{M}$ form a pair of complementary masks.

Using these two masks, we construct a pair of complementary masked images, denoted by $I^{(1)}$ and $I^{(2)}$:
\begin{equation}
I^{(1)} = I \odot M, \qquad
I^{(2)} = I \odot \bar{M},
\label{eq:masked_images}
\end{equation}
where $\odot$ denotes the patch-wise masking operation. In other words, $I^{(1)}$ preserves the patches whose indices satisfy that $(i+j)$ is even, while $I^{(2)}$ preserves the complementary set of patches.

Since the two masked images retain complementary visible regions, their reconstructed outputs are fused to obtain the final reconstructed RGB image 
$R\in\mathbb{R}^{224\times224\times3}$.

To convert the reconstructed image back to a time series, we first average the three RGB channels to obtain a reconstructed grayscale map:
\begin{equation}
R_{\mathrm{gray}}(i,j)
=
\frac{1}{3}\sum_{k=1}^{3}R(i,j,k)
\label{eq:rgb_to_gray}
\end{equation}
Then, $R_{\mathrm{gray}}\in \mathbb{R}^{224\times224}$ is resized back to the original variable-time resolution $C\times L$ through bilinear interpolation, yielding the reconstructed sequence
$\hat{X}\in\mathbb{R}^{C\times L}$.

\subsection{The Adaptive Distribution Modulation Module}
\label{section:admm}

% Directly applying MAE for reconstruction suffers from an over-generalization issue. Although $X$ and $\hat{X}$ may be close in numerical values, they often differ in their statistical characteristics. In particular, the MAE output is usually smoother and exhibits weaker local variability than the original sequence. 

To better expose the discrepancy between the original and reconstructed sequences, we design an Adaptive Distribution Modulation Module (ADMM). Rather than modifying or retraining the pre-trained MAE backbone, ADMM serves as a lightweight post-processing operation that adaptively recalibrates the MAE reconstruction according to the local statistics of the input window.
Specifically, given the original sequence $X \in \mathbb{R}^{C \times L}$, we first compute the variable-wise mean and standard deviation of $X$ as
\begin{equation}
\mu_c=\frac{1}{L}\sum_{t=1}^{L}X_{c,t}, \qquad
\sigma_c=\sqrt{\frac{1}{L}\sum_{t=1}^{L}\left(X_{c,t}-\mu_c\right)^2+\epsilon},
\label{eq:admm_orig_stats}
\end{equation}
where $\epsilon$ is a small constant for numerical stability.
Given the MAE reconstruction $\hat{X}\in\mathbb{R}^{C\times L}$, ADMM adjusts the reconstruction by
\begin{equation}
\tilde{X}_{c,t}
=
\sigma_c\hat{X}_{c,t}+\mu_c.
\label{eq:admm_mapping}
\end{equation}

In this way, ADMM makes deviations caused by abnormal patterns more distinguishable. We further provide a theoretical analysis of the reconstruction errors after ADMM and show that this post-processing operation can explicitly amplify anomaly-induced reconstruction errors.

\textbf{Theoretical Analysis}: 
This analysis is motivated by the over-generalization property of the pre-trained MAE, where anomalous patterns may still be reconstructed with small reconstruction errors.
For each variable $c$, let the original reconstruction error be
$e_{c,t}=X_{c,t}-\hat{X}_{c,t}$. After applying ADMM, the calibrated reconstruction error is given by
\begin{equation}
\begin{aligned}
\tilde{e}_{c,t}
&=
X_{c,t}-\tilde{X}_{c,t} \\
&=
X_{c,t}-(\sigma_c\hat{X}_{c,t}+\mu_c) \\
&=
X_{c,t}-\sigma_c(X_{c,t}-e_{c,t})-\mu_c \\
&=
\sigma_c e_{c,t}+(1-\sigma_c)X_{c,t}-\mu_c .
\end{aligned}
\end{equation}
For notational simplicity, we denote the remaining term as
\begin{equation}
b_{c,t}=(1-\sigma_c)X_{c,t}-\mu_c .
\end{equation}
Then the ADMM reconstruction error can be compactly written as
\begin{equation}
\tilde{e}_{c,t}=\sigma_c e_{c,t}+b_{c,t}.
\end{equation}

% Inference procedure
\begin{algorithm}[t]
  \caption{Inference procedure of VAN-AD}
  \label{alg:vanad_inference}
  \renewcommand{\algorithmicrequire}{\textbf{Input:}}
  \renewcommand{\algorithmicensure}{\textbf{Output:}}
  \begin{algorithmic}[1]
  \REQUIRE Multivariate time series window $X\in\mathbb{R}^{C\times L}$, pre-trained MAE, Normalizing Flow model $f$, trade-off parameter $\lambda$
  \ENSURE Anomaly scores $\{S_t\}_{t=1}^{L}$
  \setstretch{1.05}
  \STATE Convert $X$ into a 2D image $I$
  \STATE Generate complementary masked images $I^{(1)}$ and $I^{(2)}$ via checkerboard mask
  \FOR{$k=1$ to $2$}
      \STATE Reconstruct $I^{(k)}$ with MAE to obtain the reconstructed image $R^{(k)}$
  \ENDFOR
  \STATE Fuse $R^{(1)}$ and $R^{(2)}$ to obtain the complete reconstructed image $R$
  \STATE Resize $R$ back to the original time-series resolution to obtain $\hat{X}\in\mathbb{R}^{C\times L}$
  \STATE Compute $\mu_c$ and $\sigma_c$ for each variable $c$ of the original sequence
  \STATE Apply ADMM to obtain $\tilde{X}$, where $\tilde{X}_{c,t}=\sigma_c\hat{X}_{c,t}+\mu_c$
  \FOR{$t=1$ to $L$}
      \STATE $S_{\mathrm{MAE}}^{(t)}=\|\tilde{x}_t-x_t\|_2^2$
      \STATE $S_{\mathrm{NF}}^{(t)}=-\log p_{\hat{X}}(x_t)$
      \STATE $S_t=S_{\mathrm{MAE}}^{(t)}+\lambda S_{\mathrm{NF}}^{(t)}$
  \ENDFOR
  \STATE \textbf{return} $\{S_t\}_{t=1}^{L}$
  \end{algorithmic}
  \end{algorithm}

We then shift the point-wise residual analysis to the entire sequence. For variable $c$, the mean squared reconstruction error after ADMM is defined as
\begin{equation}
\mathrm{MSE}_{ADMM}^{(c)}
=
\frac{1}{L}\sum_{t=1}^{L}\tilde{e}_{c,t}^{2}.
\end{equation}
Substituting $\tilde{e}_{c,t}=\sigma_c e_{c,t}+b_{c,t}$, we obtain
\begin{equation}
\begin{aligned}
\mathrm{MSE}_{ADMM}^{(c)}
&=
\frac{1}{L}\sum_{t=1}^{L}
(\sigma_c e_{c,t}+b_{c,t})^2 \\
&=
\sigma_c^2 \mathrm{MSE}_{rec}^{(c)}
+
\mathrm{MSE}_{stat}^{(c)}
+
\mathrm{CROSS} \\
&\approx
\sigma_c^2 \mathrm{MSE}_{rec}^{(c)}
+
\mathrm{MSE}_{stat}^{(c)} ,
\end{aligned}
\label{eq:mse_admm}
\end{equation}
where
\begin{equation}
\mathrm{MSE}_{rec}^{(c)}
=
\frac{1}{L}\sum_{t=1}^{L}e_{c,t}^2,
\quad
\mathrm{MSE}_{stat}^{(c)}
=
\frac{1}{L}\sum_{t=1}^{L}b_{c,t}^2,
\end{equation}
and
\begin{equation}
\mathrm{CROSS}
=
2\sigma_c
\frac{1}{L}\sum_{t=1}^{L}e_{c,t}b_{c,t}.
\end{equation}
Since the pre-trained MAE backbone usually yields small reconstruction errors, i.e. $e_{c,t}\approx0$. $\mathrm{MSE}_{rec}^{(c)}$ and $\mathrm{CROSS}$ are expected to be small. We retain $\mathrm{MSE}_{rec}^{(c)}$ in Eq.~\ref{eq:mse_admm} to highlight that ADMM rescales the original reconstruction error by $\sigma_c^2$ and introduces an additional statistic-induced term $\mathrm{MSE}_{stat}^{(c)}$.

For the term $\sigma_c^2\mathrm{MSE}_{rec}^{(c)}$, ADMM explicitly rescales the original reconstruction error according to the local scale of the input window.

For $\mathrm{MSE}_{stat}^{(c)}$, we further analyze its dependence on the local statistics of the input window. According to the definition of $b_{c,t}$, and using
$\frac{1}{L}\sum_{t=1}^{L}X_{c,t}=\mu_c$ and
$\frac{1}{L}\sum_{t=1}^{L}X_{c,t}^2=\sigma_c^2+\mu_c^2$
by ignoring the small numerical constant $\epsilon$, we obtain
\begin{equation}
\begin{aligned}
\mathrm{MSE}_{stat}^{(c)}
&=
\frac{1}{L}\sum_{t=1}^{L}
\left[(1-\sigma_c)X_{c,t}-\mu_c\right]^2 \\
&=
\sigma_c^2
\left[
(1-\sigma_c)^2+\mu_c^2
\right].
\end{aligned}
\label{eq:mse_stat}
\end{equation}
This result shows that $\mathrm{MSE}_{stat}^{(c)}$ depends on both the local standard deviation $\sigma_c$ and the local mean $\mu_c$. Among them, $\sigma_c$ has a stronger influence because it appears in both the outer scaling factor $\sigma_c^2$ and the inner deviation term $(1-\sigma_c)^2$, while $\mu_c$ affects the term through $\sigma_c^2\mu_c^2$.

This analysis indicates that ADMM is particularly sensitive to local statistical changes. For normal windows, the temporal variations are usually more stable and the local standard deviation tends to remain within a normal range, so the ADMM-induced change in reconstruction error is limited. In contrast, anomalous windows often exhibit mean shifts or enlarged local variations, especially increased standard deviation, which strengthens the reconstruction errors after ADMM. In Section~\ref{sec:visualizaton of admm}, we further visualize the transformation process of normal and anomalous sequences after applying ADMM.

However, ADMM still relies on the local statistics of the current window. For long-lasting anomalies that occupy most or all of the window, the local variations may become relatively stable, making the anomaly difficult to distinguish using ADMM alone. Therefore, we further design the Normalizing Flow Module (NFM) to model the global distribution of normal patterns and complement ADMM with density-based anomaly scoring.

% dataset information
\begin{table*}[htbp]
\centering
\caption{Statistics of the datasets. All dataset normalization and train/test splits strictly follow the TAB survey~\cite{qiu2025tab}.}
\label{tab:data}
\begin{small}
\setlength{\tabcolsep}{5pt}
\begin{tabularx}{\textwidth}{@{}l l r r r  X@{}}
\toprule
\textbf{Dataset} & \textbf{Domain} & \textbf{Variable} &
\textbf{Train Length} &\textbf{Test Length} &
\textbf{Description} \\
\midrule
CalIt2 & Visitors Flowrate & 2  & 2{,}520  & 2{,}520 & Building entry/exit count time series (15 days)\\
CICIDS & Web Traffic & 72 & 85,115 & 82,116 & Network traffic data with 80+ features and attack lab\\
DLR & Health & 19 & 11{,}565 & 11{,}565 &Physiological time series with anomalies from monitoring systems\\
GECCO  & Water Treatment & 9  & 69{,}260  & 69{,}261 & Water treatment data from the GECCO Challenge\\
MSL & Spacecraft & 1 & 58{,}317 & 73{,}729 & Spacecraft telemetry and anomaly data from the MSL rover\\
NYC & Transport & 3 & 13{,}104 & 4{,}416 &Information on taxi and ride-hailing trips in New York\\
PUMP & Water Treatment & 44 & 76{,}901 & 14{,}3401 &Sensor data from pump systems in water treatment processes\\
PSM & Server Machine & 25 & 13{,}2481 & 87{,}841 & Application server telemetry data with labeled anomalies\\
SWAN & Space Weather & 38 & 60{,}000 & 60{,}000 &Space weather data (satellite and geomagnetic measurements)\\
\bottomrule
\end{tabularx}
\end{small}
\end{table*}

% % dataset information
% \begin{table*}[htbp]
%   \centering
%   \caption{Statistics of anomaly labels in the test sets.}
%   \label{tab:data1}
%   \begin{small}
%   \setlength{\tabcolsep}{5pt}
%   \begin{tabularx}{\textwidth}{@{}l l c c c  X@{}}
%   \toprule
%   \textbf{Dataset} & \textbf{Test Length} & \textbf{Anomalies} &
%   \textbf{Anomaly Ratio (\%)} & \textbf{A-P} & \textbf{V-P} \\
%   \midrule
%   CalIt2 & 2{,}520 & 74 & 2.94 & 0.108 & 0.414\\
%   CICIDS & 85{,}116 & 79 & 0.09 & 0.007 & 0.020\\
%   NYC & 4{,}416 & 99 & 2.24 & 0.151 & 0.138\\
%   MSL & 73{,}729 & 7{,}766 & 10.53 & 0.293 & 0.297\\
%   \midrule
%   PSM & 87{,}841 & 24{,}381 & 27.76 & 0.622 & 0.661\\
%   PUMP & 143{,}401 & 14{,}408 & 10.05 & 0.559 & 0.538\\
%   SWAN & 60{,}000 & 19{,}560 & 32.60 & 0.687 & 0.924\\
%   \bottomrule
%   \end{tabularx}
%   \end{small}
%   \end{table*}

% Main Results
\begin{table*}[htbp]
  \centering
  \small
  \caption{A-R (AUC-ROC), A-P (AUC-PR), V-R (VUS-ROC), and V-P (VUS-PR) results on nine real-world datasets, where higher values indicate better performance. \textbf{Ours} denotes VAN-AD in the full-shot setting, while \textbf{Ours (ZS)} denotes the zero-shot variant obtained by removing NFM from VAN-AD. The best results are highlighted in bold, and the second-best results are underlined.
  }
  \setlength{\tabcolsep}{2pt}
  \renewcommand{\arraystretch}{1.35}
  \resizebox{\textwidth}{!}{%
  \begin{tabular}{>{\centering\arraybackslash}p{1.3cm}|c|*{16}{>{\centering\arraybackslash}p{1.30cm}}}
  \hline
  Dataset & Metric & \textbf{Ours} & \textbf{Ours(ZS)} & DADA & Timer & CALF & GPT4TS & CATCH & MtsCID & CAROTS & iTrans & TFMAE & MMA & LOF & CBLOF & KNN & PCA\\
  \hline
  
  \multirow{4}{*}{CaIIt2}
   & A-R & \textbf{0.797} & 0.788 & 0.759 & 0.739 & 0.614 & 0.731 & 0.772 & 0.610 & 0.767 & \underline{0.791} & 0.351 & 0.779 & 0.509 & 0.772 & 0.539 & 0.790\\
   & A-P & 0.108 & 0.108 & 0.090 & 0.079 & 0.052 & 0.082 & 0.089 & 0.048 & \textbf{0.188} & 0.106 & 0.026 & \underline{0.137} & 0.035 & 0.093 & 0.037 & 0.073\\
   & V-R & \textbf{0.953} & 0.947 & 0.796 & 0.785 & 0.668 & 0.766 & 0.796 & 0.643 & 0.903 & 0.809 & 0.766 & \underline{0.949} & 0.532 & 0.808 & 0.514 & 0.786\\
   & V-P & \textbf{0.414} & \underline{0.412} & 0.096 & 0.096 & 0.060 & 0.087 & 0.101 & 0.326 & 0.375 & 0.110 & 0.158 & 0.342 & 0.061 & 0.111 & 0.101 & 0.103\\
  \hline
  
  \multirow{4}{*}{CICIDS}
   & A-R & 0.700 & 0.603 & 0.682 & 0.679 & 0.617 & \underline{0.712} & 0.669 & 0.500 & 0.580 & 0.668 & 0.542 & 0.674 & 0.458 & \textbf{0.804} & 0.564 & 0.601\\
   & A-P & \textbf{0.007} & 0.001 & 0.001 & 0.001 & 0.001 & \underline{0.002} & 0.001 & 0.001 & \textbf{0.007} & 0.001 & 0.001 & 0.001 & 0.001 & \underline{0.002} & 0.001 & 0.001\\
   & V-R & \textbf{0.926} & 0.883 & 0.641 & 0.647 & 0.518 & 0.660 & 0.598 & 0.507 & 0.722 & 0.578 & 0.848 & \underline{0.916} & 0.332 & 0.773 & 0.465 & 0.496\\
   & V-P & \textbf{0.020} & 0.014 & 0.002 & 0.002 & 0.001 & 0.002 & 0.002 & \textbf{0.020} & 0.015 & 0.002 & \underline{0.016} & 0.015 & 0.001 & 0.003 & 0.001 & 0.001\\
  \hline
  
  \multirow{4}{*}{DLR}
   & A-R & \textbf{0.980} & \underline{0.979} & 0.948 & 0.907 & 0.900 & 0.902 & 0.922 & 0.497 & 0.744 & 0.951 & 0.186 & 0.975 & 0.396 & 0.880 & 0.889 & 0.911\\
   & A-P & \textbf{0.835} & \textbf{0.835} & 0.170 & 0.238 & 0.193 & 0.108 & 0.171 & 0.015 & 0.035 & 0.194 & 0.009 & 0.273 & 0.007 & 0.075 & 0.070 & \underline{0.295}\\
   & V-R & \textbf{0.994} & \underline{0.993} & 0.940 & 0.939 & 0.930 & 0.899 & 0.921 & 0.496 & 0.785 & 0.933 & 0.226 & 0.978 & 0.473 & 0.874 & 0.885 & 0.960\\
   & V-P & \textbf{0.821} & \underline{0.820} & 0.190 & 0.357 & 0.175 & 0.091 & 0.161 & 0.013 & 0.041 & 0.169 & 0.013 & 0.264 & 0.060 & 0.071 & 0.072 & 0.329\\
  \hline
  
  \multirow{4}{*}{GECCO}
   & A-R & \underline{0.978} & \textbf{0.985} & 0.772 & 0.955 & 0.866 & 0.836 & 0.912 & 0.535 & 0.507 & 0.780 & 0.315 & 0.892 & 0.796 & 0.684 & 0.811 & 0.711\\
   & A-P & \textbf{0.619} & \underline{0.573} & 0.264 & 0.482 & 0.189 & 0.162 & 0.198 & 0.019 & 0.147 & 0.081 & 0.009 & 0.481 & 0.086 & 0.072 & 0.170 & 0.234\\
   & V-R & \textbf{0.996} & \textbf{0.996} & 0.706 & 0.935 & 0.930 & 0.908 & \underline{0.953} & 0.514 & 0.477 & 0.852 & 0.548 & 0.943 & 0.767 & 0.678 & 0.756 & 0.595\\
   & V-P & \textbf{0.722} & \underline{0.706} & 0.046 & 0.305 & 0.235 & 0.256 & 0.260 & 0.038 & 0.038 & 0.107 & 0.019 & 0.363 & 0.058 & 0.038 & 0.068 & 0.046\\
  \hline
  
  \multirow{4}{*}{MSL}
   & A-R & \textbf{0.762} & \underline{0.755} & 0.547 & 0.637 & 0.592 & 0.594 & 0.627 & 0.500 & 0.557 & 0.589 & 0.488 & 0.550 & 0.557 & 0.622 & 0.623 & 0.552\\
   & A-P & \textbf{0.293} & \underline{0.277} & 0.138 & 0.155 & 0.143 & 0.137 & 0.154 & 0.106 & 0.154 & 0.141 & 0.099 & 0.141 & 0.120 & 0.181 & 0.200 & 0.157\\
   & V-R & \textbf{0.778} & \underline{0.773} & 0.630 & 0.700 & 0.669 & 0.665 & 0.699 & 0.511 & 0.612 & 0.668 & 0.527 & 0.613 & 0.617 & 0.684 & 0.642 & 0.622\\
   & V-P & \textbf{0.297} & \underline{0.288} & 0.200 & 0.229 & 0.215 & 0.199 & 0.229 & 0.150 & 0.167 & 0.217 & 0.116 & 0.162 & 0.181 & 0.224 & 0.242 & 0.200\\
  \hline
  
  \multirow{4}{*}{NYC}
   & A-R & \textbf{0.881} & \textbf{0.881} & 0.659 & 0.661 & 0.457 & 0.495 & 0.469 & 0.515 & 0.519 & 0.594 & 0.410 & 0.544 & 0.464 & 0.528 & 0.466 & \underline{0.666}\\
   & A-P & \textbf{0.151} & \textbf{0.151} & 0.035 & 0.034 & 0.020 & 0.027 & 0.021 & 0.023 & 0.026 & 0.030 & 0.020 & 0.024 & 0.020 & 0.024 & 0.020 & \underline{0.042}\\
   & V-R & \textbf{0.894} & \textbf{0.894} & 0.689 & 0.704 & 0.573 & 0.628 & 0.593 & 0.530 & 0.622 & 0.663 & 0.535 & 0.637 & 0.581 & 0.641 & 0.608 & \underline{0.730}\\
   & V-P & \underline{0.138} & \textbf{0.139} & 0.051 & 0.054 & 0.042 & 0.052 & 0.040 & 0.043 & 0.039 & 0.048 & 0.031 & 0.032 & 0.049 & 0.052 & 0.051 & 0.072\\
  \hline
  
  \multirow{4}{*}{PUMP}
   & A-R & \textbf{0.937} & 0.466 & 0.788 & 0.529 & 0.523 & 0.414 & 0.466 & 0.500 & 0.785 & 0.577 & 0.400 & \underline{0.805} & 0.479 & 0.621 & 0.524 & 0.802\\
   & A-P & \textbf{0.559} & 0.134 & 0.199 & 0.139 & 0.126 & 0.101 & 0.104 & 0.102 & 0.210 & 0.123 & 0.084 & \underline{0.213} & 0.089 & 0.118 & 0.104 & 0.209\\
   & V-R & \textbf{0.921} & 0.479 & 0.789 & 0.712 & 0.644 & 0.551 & 0.583 & 0.500 & 0.779 & 0.668 & 0.379 & \underline{0.803} & 0.537 & 0.605 & 0.536 & 0.793\\
   & V-P & \textbf{0.538} & 0.142 & 0.228 & \underline{0.260} & 0.210 & 0.177 & 0.168 & 0.107 & 0.208 & 0.190 & 0.086 & 0.224 & 0.127 & 0.155 & 0.138 & 0.248\\
  \hline
  
  \multirow{4}{*}{PSM}
   & A-R & \textbf{0.815} & 0.717 & 0.621 & 0.556 & 0.583 & 0.584 & 0.600 & 0.500 & \underline{0.767} & 0.583 & 0.333 & 0.636 & 0.730 & 0.717 & 0.744 & 0.667\\
   & A-P & \textbf{0.622} & 0.527 & 0.448 & 0.359 & 0.369 & 0.380 & 0.385 & 0.280 & 0.537 & 0.372 & 0.215 & 0.435 & 0.462 & 0.529 & \underline{0.553} & 0.477\\
   & V-R & \textbf{0.817} & 0.745 & 0.599 & 0.541 & 0.579 & 0.579 & 0.600 & 0.515 & \underline{0.750} & 0.582 & 0.352 & 0.667 & 0.641 & 0.636 & 0.667 & 0.576\\
   & V-P & \textbf{0.661} & \underline{0.585} & 0.423 & 0.350 & 0.370 & 0.379 & 0.389 & 0.320 & 0.529 & 0.376 & 0.235 & 0.453 & 0.423 & 0.442 & 0.456 & 0.413\\
  \hline
  
  \multirow{4}{*}{SWAN}
   & A-R & 0.829 & 0.701 & 0.645 & 0.624 & 0.515 & 0.503 & 0.485 & 0.500 & 0.613 & 0.512 & 0.561 & 0.716 & 0.740 & \textbf{0.889} & \underline{0.878} & 0.666\\
   & A-P & 0.687 & 0.597 & 0.550 & 0.511 & 0.411 & 0.454 & 0.425 & 0.329 & \textbf{0.846} & 0.419 & 0.140 & 0.580 & 0.598 & \underline{0.770} & 0.764 & 0.544\\
   & V-R & \textbf{0.964} & \underline{0.949} & 0.561 & 0.547 & 0.425 & 0.447 & 0.415 & 0.803 & 0.807 & 0.435 & 0.568 & 0.942 & 0.601 & 0.744 & 0.798 & 0.483\\
   & V-P & \textbf{0.924} & \underline{0.918} & 0.487 & 0.461 & 0.376 & 0.407 & 0.376 & 0.838 & 0.904 & 0.382 & 0.143 & 0.888 & 0.517 & 0.711 & 0.696 & 0.448\\
  \hline
  \multirow{4}{*}{\textbf{Average}}
   & A-R & \textbf{0.853} & \underline{0.764} & 0.713 & 0.698 & 0.630 & 0.641 & 0.658 & 0.517 & 0.649 & 0.672 & 0.398 & 0.730 & 0.570 & 0.724 & 0.671 & 0.707\\
   & A-P & \textbf{0.431} & \underline{0.356} & 0.255 & 0.228 & 0.178 & 0.183 & 0.192 & 0.123 & 0.329 & 0.170 & 0.067 & 0.254 & 0.159 & 0.239 & 0.213 & 0.227\\
   & V-R & \textbf{0.916} & \underline{0.851} & 0.705 & 0.699 & 0.659 & 0.678 & 0.684 & 0.554 & 0.736 & 0.659 & 0.528 & 0.828 & 0.612 & 0.742 & 0.652 & 0.669\\
   & V-P & \textbf{0.504} & \underline{0.447} & 0.253 & 0.260 & 0.209 & 0.227 & 0.231 & 0.327 & 0.405 & 0.222 & 0.091 & 0.305 & 0.205 & 0.248 & 0.203 & 0.206\\
  \hline
  \end{tabular}
  }
  
  \label{tab:main-results}
  \end{table*}

\subsection{The Normalizing Flow Module}
Compared with the original sequence, the reconstructed sequence tends to suppress noise and retain more regular temporal patterns. This property makes it a more suitable basis for modeling the distribution of normal patterns. Based on this observation, we employ a normalizing flow on top of the reconstructed sequence to model the global probability density of the normal patterns.

For each reconstructed observation $\hat{x}_t \in \mathbb{R}^{C}$, a normalizing flow
$f:\mathbb{R}^{C}\rightarrow\mathbb{R}^{C}$ is used to model the corresponding density
$p_{\hat{X}}(\hat{x}_t)$. Specifically, $\hat{x}_t$ is mapped to a latent variable
$z_t=f(\hat{x}_t)$. The base density is defined as a Gaussian distribution with mean vector
$u \in \mathbb{R}^{C}$ and identity covariance, i.e.,
$p_Z(z_t)=\mathcal{N}(z_t \mid u,I)$,
where $u$ is randomly initialized once at model initialization and then kept fixed.

Inspired by MAF~\cite{papamakarios2017masked}, we adopt Masked Autoregressive Flow as the normalizing flow architecture. As illustrated in Fig. \ref{fig:van-ad}, the flow $f$ consists of $N$ stacked layers, each composed of a Masked Linear transformation followed by an activation function. To capture dependencies among variables more effectively, the causal dependency matrix is reversed at each layer, so that every variable can progressively interact with all other variables throughout the flow.

Consider an autoregressive model whose conditional distributions are parameterized as univariate Gaussians. The $i$-th conditional distribution is defined as
\begin{equation}
p\left(\hat{x}_{t,i}\mid \hat{x}_{t,1:i-1}\right)
=
\mathcal{N}\left(\hat{x}_{t,i}\mid \mu_i,\left(\exp \alpha_i\right)^2\right),
\label{eq:maf_conditional}
\end{equation}
where $\mu_i=f_{\mu_i}\left(\hat{x}_{t,1:i-1}\right)$ and
$\alpha_i=f_{\alpha_i}\left(\hat{x}_{t,1:i-1}\right)$.

After estimating $\mu_i$ and $\alpha_i$, the transformation from $\hat{x}_t$ to $z_t$ is given by
\begin{equation}
z_{t,i}
=
\bigl[f(\hat{x}_t)\bigr]_i
=
\left(\hat{x}_{t,i}-\mu_i\right)\exp\left(-\alpha_i\right).
\label{eq:maf_transform}
\end{equation}

Due to the autoregressive structure, the Jacobian matrix of $f$ is triangular. Therefore, its absolute determinant can be efficiently computed as
\begin{equation}
\left|\det \nabla_{\hat{x}_t} f(\hat{x}_t)\right|
=
\exp\left(-\sum_{i=1}^{C}\alpha_i\right).
\label{eq:maf_jacobian}
\end{equation}
Accordingly, the log density of $\hat{x}_t$ can be written as
\begin{equation}
\begin{aligned}
\log p_{\hat{X}}(\hat{x}_t)
&=
\log p_Z\bigl(f(\hat{x}_t)\bigr)
+
\log \left|\det \nabla_{\hat{x}_t} f(\hat{x}_t)\right| \\
&=
-\frac{1}{2}\sum_{i=1}^{C}\left[\left(z_{t,i}-u_i\right)^2+\log 2\pi\right]
-\sum_{i=1}^{C}\alpha_i.
\end{aligned}
\label{eq:maf_log_density}
\end{equation}

\subsection{Training Objective}

The pre-trained MAE backbone is kept frozen throughout VAN-AD, and ADMM is a training-free post-processing operation. 
Thus, MAE with ADMM can be directly applied to different target datasets without updating the visual backbone or introducing extra trainable parameters.

When target-domain normal data are available, NFM is introduced as a lightweight adaptation module to model dataset-specific normal distributions. 
During this stage, only the normalizing flow is optimized, while the MAE backbone remains frozen. 
The flow is trained by maximizing the likelihood of normal data, which is equivalent to minimizing the Kullback--Leibler divergence between the true data distribution and the flow-induced distribution. Accordingly, the training objective is defined as
\begin{equation}
\min_{\theta}\mathcal{L}(\theta,\hat{X})
=
\frac{1}{T}\sum_{t=1}^{T}-\log p_{\hat{X}}(\hat{x}_t),
\label{eq:nfm_loss}
\end{equation}
where $\theta$ denotes the parameters of the normalizing flow.

\subsection{Anomaly Detection}

Based on the assumption that anomalous data are difficult to reconstruct, we use the reconstruction discrepancy before and after MAE to quantify the anomaly score at each time step. In addition, since anomalies usually deviate from the majority of data instances, they are also expected to have lower probability density under the learned normal pattern distribution. Therefore, we further incorporate the density estimated by the normalizing flow as an additional anomaly indicator, where lower density implies a higher likelihood of anomaly.

Specifically, the original observation $x_t$ is evaluated under the flow-based distribution learned from the reconstructed sequence. The anomaly score at time step $t$ is defined as
\begin{equation}
S_t = S_{\mathrm{REC}}^{(t)} + \lambda S_{\mathrm{NF}}^{(t)},
\label{eq:final_score}
\end{equation}
where
\begin{equation}
S_{\mathrm{REC}}^{(t)} = \left\| \tilde{x}_t - x_t \right\|_2^2,
\label{eq:mae_score}
\end{equation}
\begin{equation}
S_{\mathrm{NF}}^{(t)} = -\log p_{\hat{X}}(x_t),
\label{eq:nf_score}
\end{equation}
and $\lambda$ controls the contribution of the density based anomaly score estimated by the normalizing flow. The detailed inference procedure is presented in Algorithm~\ref{alg:vanad_inference}.

\section{Experiments}

\subsection{Datasets}
We evaluate VAN-AD on nine real-world datasets: CalIt2, CICIDS, DLR, GECCO, MSL, NYC, PUMP, PSM, and SWAN. These datasets are widely used in TSAD and cover a diverse range of application domains. Detailed dataset statistics are reported in Table \ref{tab:data}. All dataset normalization and train/test splits strictly follow the settings summarized in the TAB survey~\cite{qiu2025tab}. The same normalization strategy and data splits are applied to all baselines to ensure a fair comparison.

\subsection{Baselines} 
We compare VAN-AD with 14 baseline methods, including TS-based methods, Text-based methods, recent deep learning baselines, and classical anomaly detection methods. 
Specifically, the TS-based methods include DADA~\cite{shentu2025towards} and Timer~\cite{liu2024timer}, while the Text-based methods include CALF~\cite{liu2025calf} and GPT4TS~\cite{zhou2023one}. 
All these methods are evaluated under the full-shot setting. We also include several recent deep learning baselines, including CATCH~\cite{wucatch}, MtsCID~\cite{xie2025multivariate}, CAROTS~\cite{kim2025causality}, iTransformer (iTrans)~\cite{liuitransformer}, TFMAE~\cite{fang2024temporal}, and MMA~\cite{tang2024mlp}. In addition, we consider several classical anomaly detection methods following the TAB settings~\cite{qiu2025tab}, including Local Outlier Factor (LOF)~\cite{breunig2000lof}, Cluster-based Local Outlier Factor (CBLOF)~\cite{he2003discovering}, K-th Nearest Neighbor (KNN)~\cite{ramaswamy2000efficient}, and Principal Component Analysis (PCA)~\cite{shyu2003novel}. LOF, CBLOF, and KNN are implemented point-wise, whereas PCA uses sliding-window features with its window score assigned to the starting timestamp.

% Abaltion Study
\begin{table*}
  \caption{Ablation studies for VAN-AD. The A-R, A-P, V-R, and V-P denote AUC-ROC, AUC-PR, VUS-ROC, and VUS-PR, respectively, where higher values indicate better performance. Specifically, the origin setting directly applies the original MAE for inference. The best ones are highlighted in bold. 
  }
  \centering
  \small
  \setlength{\tabcolsep}{4pt}
  \renewcommand{\arraystretch}{1.7}
  \resizebox{\textwidth}{!}{%
    \begin{tabular}{ll|cccc|cccc|cccc|cccc|cccc}
    \hline
    \multicolumn{2}{l|}{Dataset} & \multicolumn{4}{c|}{CalIt2} & \multicolumn{4}{c|}{MSL} & \multicolumn{4}{c|}{PSM} & \multicolumn{4}{c|}{SWAN} & \multicolumn{4}{c}{PUMP}\\
    \hline
    \multicolumn{2}{l|}{Metric} & A-R & A-P & V-R & V-P & A-R & A-P & V-R & V-P & A-R & A-P & V-R & V-P & A-R & A-P & V-R & V-P & A-R & A-P & V-R & V-P\\
    \hline
    \multicolumn{2}{l|}{Origin} & 0.759 & 0.068 & 0.933 & 0.290 & 0.610 & 0.206 & 0.672 & 0.223 & 0.635 & 0.494 & 0.655 & 0.480 & 0.524 & 0.483 & 0.939 & 0.896 & 0.809 & 0.216 & 0.806 & 0.231\\
    \hline
    \multicolumn{2}{l|}{w/o ADMM} & 0.794 & 0.089 & 0.945 & 0.326 & 0.588 & 0.203 & 0.645 & 0.217 & 0.766 & 0.581 & 0.763 & 0.562 & 0.827 & 0.684 & 0.963 & 0.920 & 0.935 & 0.557 & 0.917 & 0.533\\
    \hline
    \multicolumn{2}{l|}{w/o NF} & 0.788 & \textbf{0.108} & 0.947 & 0.412 & 0.755 & 0.277 & 0.773 & 0.288 & 0.717 & 0.527 & 0.745 & 0.585 & 0.701 & 0.597 & 0.949 & 0.918 & 0.466 & 0.134 & 0.479 & 0.142\\
    \hline
    \multicolumn{2}{l|}{VAN-AD} & \textbf{0.797} & \textbf{0.108} & \textbf{0.953} & \textbf{0.414} & \textbf{0.762} & \textbf{0.293} & \textbf{0.778} & \textbf{0.297} & \textbf{0.815} & \textbf{0.622} & \textbf{0.817} & \textbf{0.661} & \textbf{0.829} & \textbf{0.687} & \textbf{0.964} & \textbf{0.924} & \textbf{0.937} & \textbf{0.559} & \textbf{0.921} & \textbf{0.538}\\
    \hline
    \end{tabular}
  }
  \label{tab:abla}
  \end{table*}

% Backbone Analysis
\begin{figure}[t]
  \centering
  \includegraphics[width=1.0\linewidth]{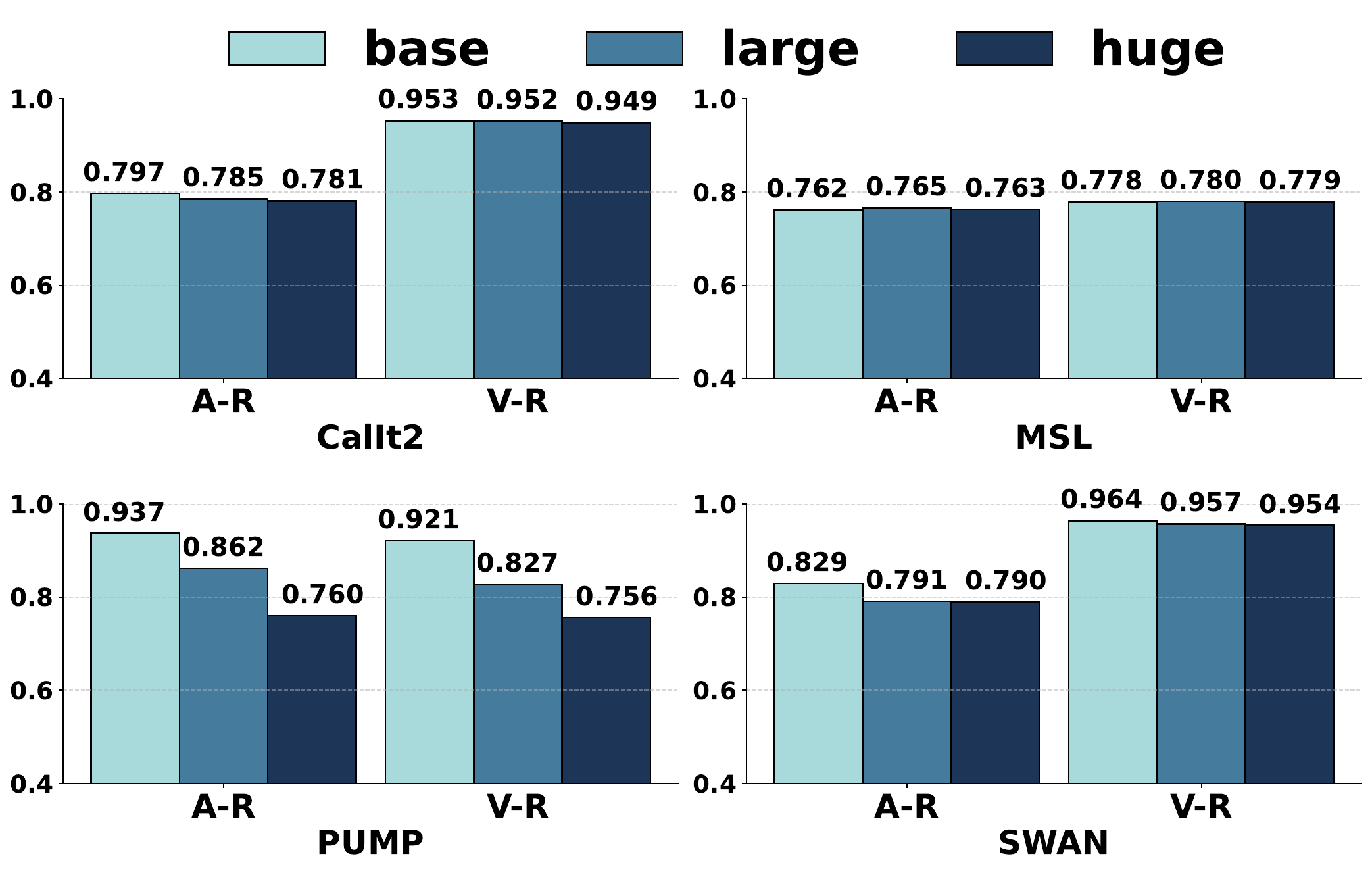}
  \caption{Backbone analysis of MAE variants with different model sizes (base, large, and huge), evaluated by A-R and V-R.}
  \label{fig:backbone_overview}
\end{figure}

% Imaging Experiments
\begin{table}
  \centering
  \small
  \caption{Imaging experiments for VAN-AD, evaluated by V-R and V-P. And time indicates the duration required to process the entire test dataset. The best ones are in bold.}
  \setlength{\tabcolsep}{4pt}
  \renewcommand{\arraystretch}{1.7}
  \resizebox{\columnwidth}{!}{%
    \begin{tabular}{ll|ccc|ccc|ccc}
    \hline
    \multicolumn{2}{l|}{Dataset} & \multicolumn{3}{c|}{PSM} & \multicolumn{3}{c|}{PUMP} & \multicolumn{3}{c}{SWAN}\\
    \hline
    \multicolumn{2}{l|}{Metric} & V-R & V-P & Time(s) & V-R & V-P & Time(s) & V-R & V-P & Time(s)\\
    \hline
    \multicolumn{2}{l|}{Seg} & 0.801 & 0.651 & 37.51 & 0.802 & 0.224 & 210.29 & 0.961 & 0.918 & 38.97\\
    \hline
    \multicolumn{2}{l|}{GAF} & 0.804 & 0.648 & 37.42 & \textbf{0.956} & 0.535 & 209.67 & 0.958 & \textbf{0.926} & 38.99\\
    \hline
    \multicolumn{2}{l|}{STFT} & 0.621 & 0.384 & 37.30 & 0.807 & 0.222 & 209.55 & 0.943 & 0.891 & 39.01\\
    \hline
    \multicolumn{2}{l|}{Wavelet} & 0.528 & 0.322 & 37.58 & 0.768 & 0.189 & 210.75 & \textbf{0.965} & 0.920 & 39.51\\
    \hline
    \multicolumn{2}{l|}{RP} & 0.561 & 0.332 & 37.30 & 0.520 & 0.098 & 209.61 & 0.951 & 0.923 & 38.99\\
    \hline
    \multicolumn{2}{l|}{Heatmap} & \textbf{0.817} & \textbf{0.661} & \textbf{2.49} & 0.921 & \textbf{0.538} & \textbf{5.07} & 0.964 & 0.924 & \textbf{1.17}\\
    \hline
    \end{tabular}
  }
  \label{tab:imaging}
  \end{table}
  
\subsection{Setup}
\subsubsection{Metrics}
Recent studies have shown that the point adjustment strategy widely used in time series anomaly detection may produce overly optimistic results, and can even yield state-of-the-art performance when applied to random methods\cite{wangd3r}. To address this issue, some studies have adopted Affiliation-F1\cite{huet2022local}, which measures recall by considering the average directional distance between predicted events and ground-truth events. However, TSB-AD\cite{liu2024elephant} points out that Affiliation-F1 has limited discriminative ability and cannot effectively reflect differences in prediction quality. Therefore, we adopt threshold-independent evaluation metrics, including the Area Under the Receiver Operating Characteristic Curve (AUC-ROC), Area Under the Precision-Recall Curve (AUC-PR), Volume Under the Surface-Receiver Operating Characteristic (VUS-ROC), and Volume Under the Surface-Precision-Recall (VUS-PR).
\subsubsection{Implementation Details}
In our experiments, we use the MAE-base model (112M) as the visual backbone. 
For the Normalizing Flow Module, the default batch size is set to 128 and is reduced by half when out-of-memory (OOM) errors occur. 
We use $\tanh$ as the activation function because its smooth and bounded nature helps maintain numerical stability, and set the number of flow layers $N$ to 3 by default. 

For each dataset, we follow the train/test split reported in Table~\ref{tab:data}. 
Following the standard unsupervised TSAD setting, the training split is used for model fitting and is assumed to contain normal data only, while the labeled test split is used only for final evaluation. 
We adopt a fixed-budget training strategy, where all models are trained for 5 epochs with a learning rate of 0.005, without using any validation set for early stopping, model selection, or hyperparameter tuning. 
All experiments are conducted on a single NVIDIA A800 GPU with 40 GB of memory.

\subsection{Main Results}
We evaluate two variants of VAN-AD against 14 competitive baselines on nine real-world datasets, as reported in Table~\ref{tab:main-results}. \textbf{Ours} denotes the full model with NFM trained on target-domain normal data, while \textit{Ours (ZS)} removes NFM and retains only the frozen MAE with ADMM for zero-shot inference. In the zero-shot setting, \textbf{Ours (ZS)} achieves average A-R, A-P, V-R, and V-P scores of 0.764, 0.356, 0.851, and 0.447, respectively, demonstrating the transferability of the visual backbone to TSAD. With NFM-based target-domain adaptation, \textit{Ours} further improves these scores to 0.853, 0.431, 0.916, and 0.504, respectively. Compared with the representative TS-based foundation model DADA~\cite{shentu2025towards}, VAN-AD improves the average A-R, A-P, V-R, and V-P scores by 14.0\%, 17.6\%, 21.1\%, and 25.1\%, respectively, with particularly large gains in V-R and V-P under range-aware evaluation.

Despite the overall strong performance, the A-P and V-P scores remain relatively low on CalIt2, CICIDS, MSL, and NYC. 
CalIt2, CICIDS, and NYC suffer from severe class imbalance, where anomalous points occupy only a very small portion of the test set. 
In contrast, MSL contains more anomalous points, but its anomalies are often short and closely distributed, which makes precise point-level ranking difficult. 
Similar trends are observed for most baselines, suggesting that PR-based evaluation remains challenging under severe imbalance or densely distributed short anomalies. In future work, we plan to explore anomaly generation techniques to enrich rare anomalous patterns and alleviate these limitations.

\subsection{Ablation Study}
As shown in Table \ref{tab:abla}, removing either ADMM or NFM consistently degrades the performance of VAN-AD. For ADMM, on PSM, VAN-AD improves A-P from 0.581 to 0.622 and V-P from 0.562 to 0.661 compared with the variant without ADMM, confirming that ADMM helps mitigate the over-generalization of the pre-trained MAE and preserves reconstruction discrepancies for anomaly detection. For NFM, the improvement is more evident on PUMP, where V-R and V-P increase from 0.479 and 0.142 to 0.921 and 0.538, respectively. This indicates that global temporal context is important for distinguishing long-span or locally normal but globally anomalous patterns.

Directly applying the original MAE for inference yields inferior performance compared with the complete VAN-AD framework. Combined with the results in Table \ref{tab:main-results}, this indicates that, although image data and time series data share certain structural similarities, a model pre-trained on image data still requires task-specific adaptation for TSAD. These results further demonstrate the effectiveness of the proposed modules in adapting MAE to TSAD.

\subsection{Further Analysis of MAE for TSAD}

\subsubsection{Backbone Analysis}
Fig. \ref{fig:backbone_overview} compares the anomaly detection performance of three MAE variants with different parameter scales (112M, 330M, and 657M). It can be observed that increasing the model size does not lead to better TSAD performance, and even results in a slight decline. A possible reason is that larger vision models tend to overfit image-specific features, thereby reducing their transferability, which is consistent with the findings reported in VisionTS\cite{chen2025visionts}.

\subsubsection{Imaging Analysis}
Existing studies summarize six mainstream encoding strategies for time series representation\cite{ni2025harnessing}, namely Segment (Seg), Gramian Angular Field (GAF), STFT, Wavelet, Recurrence Plot (RP), and Heatmap. 
% Detailed descriptions of the six encoding strategies are provided in Table \ref{tab:encoding_compare}.

The experimental results for different encoding strategies are shown in Table \ref{tab:imaging}. Although heatmap encoding is not consistently the best across all datasets, it achieves competitive V-R and V-P performance with substantially lower inference time. This is because, except for Heatmap, the other methods are primarily designed for univariate time series and therefore require channel-independent inference. As a result, the computational cost increases significantly when the number of variables is large, as in datasets such as PSM, PUMP, and SWAN. Considering the importance of inference efficiency in real-world applications, we adopt Heatmap because it is naturally suited to multivariate time series encoding and achieves a better trade-off between detection performance and computational cost.

% NF Analysis
% \begin{figure}[t]
%   \centering
%   \includegraphics[width=1.0\linewidth]{figures/nf_combined_results.pdf}
%   \caption{Density modeling analysis evaluated by A-R and V-R. The compared methods include VAN-AD, MTGFLOW, and GANF.}
%   \label{fig:nf_experiment}
% \end{figure}

% Interpolation Analysis
\begin{table}
  \caption{Performance comparison of different interpolation strategies in heatmap resizing. The best ones are in bold.}
  \centering
  \small
  \setlength{\tabcolsep}{4pt}
  \renewcommand{\arraystretch}{1.5}
  \resizebox{\columnwidth}{!}{%
    \begin{tabular}{ll|cc|cc|cc|cc}
    \hline
    \multicolumn{2}{l|}{Dataset} & \multicolumn{2}{c|}{CalIt2} & \multicolumn{2}{c|}{PSM} & \multicolumn{2}{c|}{PUMP} & \multicolumn{2}{c}{SWAN}\\
    \hline
    \multicolumn{2}{l|}{Metric} & V-R & V-P & V-R & V-P & V-R & V-P & V-R & V-P\\
    \hline
    \multicolumn{2}{l|}{Nearest} & 0.943 & 0.364 & 0.687 & 0.486 & 0.705 & 0.168 & 0.951 & 0.901\\
    \hline
    \multicolumn{2}{l|}{Bicubic} & 0.951 & \textbf{0.417} & 0.814 & 0.659 & \textbf{0.941} & \textbf{0.542} & \textbf{0.964} & 0.922\\
    \hline
    \multicolumn{2}{l|}{Bilinear} & \textbf{0.953} & 0.414 & \textbf{0.817} & \textbf{0.661} & 0.921 & 0.538 & \textbf{0.964} & \textbf{0.924} \\
    \hline
  
    \end{tabular}
  }
  \label{tab:interpolation_analysis}
\end{table}

% Interpolation Similarity Analysis
\begin{table}
  \caption{Interpolation Similarity Analysis. $\uparrow$ indicates that larger values denote higher similarity, while $\downarrow$ indicates that smaller values denote higher similarity.}
  \centering
  \small
  \setlength{\tabcolsep}{4pt}
  \renewcommand{\arraystretch}{1.5}
  \resizebox{\columnwidth}{!}{%
    \begin{tabular}{ll|ccc|ccc}
    \hline
    \multicolumn{2}{l|}{Category} & \multicolumn{3}{c|}{Temporal} & \multicolumn{3}{c}{Variable}\\
    \hline
    \multicolumn{2}{l|}{Dataset} & TC$\uparrow$ & NMSE$\downarrow$ & FTC$\uparrow$ & CMS$\uparrow$ & RCME$\downarrow$ & TNPR$\uparrow$\\
    \hline
    \multicolumn{2}{l|}{CICIDS} & 0.722 & 0.117 & 0.700 & 0.718 & 0.744 & 0.590\\
    \hline
    \multicolumn{2}{l|}{DLR} & 0.998 & 0.0004 & 0.993 & 0.998 & 0.045 & 0.969\\
    \hline
    \multicolumn{2}{l|}{GECCO} & 0.904 & 0.0003 & 0.904 & 0.817 & 0.447 & 0.842\\
    \hline
    \multicolumn{2}{l|}{NYC} & 0.999 & 0.001 & 0.992 & 1.000 & 0.012 & 1.000\\
    \hline
    \multicolumn{2}{l|}{PSM} & 0.985 & 0.005 & 0.902 & 0.965 & 0.195 & 0.889\\
    \hline
    \multicolumn{2}{l|}{PUMP} & 0.961 & 0.004 & 0.959 & 0.945 & 0.287 & 0.812\\
    \hline
    \multicolumn{2}{l|}{SWAN} & 0.756 & 0.121 & 0.730 & 0.894 & 0.431 & 0.792\\
    \hline
  
    \end{tabular}
  }
  \label{tab:interpolation_similarity}
\end{table}

% Finetune Study
\begin{table*}
  \centering
  \small
  \setlength{\tabcolsep}{4pt}
  \renewcommand{\arraystretch}{1.7}
  \caption{Finetune experiments for VAN-AD.The best ones are in bold.}
  \resizebox{\linewidth}{!}{%
    \begin{tabular}{ll|cccc|cccc|cccc|cccc|cccc}
    \hline
    \multicolumn{2}{l|}{Dataset} & \multicolumn{4}{c|}{CalIt2} & \multicolumn{4}{c|}{MSL} & \multicolumn{4}{c|}{PSM} & \multicolumn{4}{c|}{PUMP} & \multicolumn{4}{c}{SWAN}\\
    \hline
    \multicolumn{2}{l|}{Metric} & A-R & A-P & V-R & V-P & A-R & A-P & V-R & V-P & A-R & A-P & V-R & V-P& A-R & A-P & V-R & V-P& A-R & A-P & V-R & V-P\\
    \hline
    \multicolumn{2}{l|}{ln} & 0.779 & 0.097 & 0.952 & \textbf{0.417} & \textbf{0.767} & 0.262 & \textbf{0.811} & 0.294
    & 0.655 & 0.378 & 0.650 & 0.396 & 0.596 & 0.110 & 0.596 & 0.113 & 0.626 & 0.480 & 0.931 & 0.868\\
    \hline
    \multicolumn{2}{l|}{bias} & 0.763 & 0.096 & 0.952 & \textbf{0.417} & 0.679 & 0.225 & 0.736 & 0.252
    & 0.626 & 0.347 & 0.629 & 0.370 & 0.782 & 0.185 & 0.779 & 0.200 & 0.450 & 0.333 & 0.900 & 0.823\\
    \hline
    \multicolumn{2}{l|}{attn} & 0.760 &0.094 &0.946 &0.345 &0.685 &0.225 &0.729 & 0.258 & 0.664 & 0.382 & 0.637 & 0.401 & 0.485 & 0.092 & 0.491 & 0.092 & 0.566 & 0.440 & 0.925 & 0.852\\
    \hline
    \multicolumn{2}{l|}{mlp} & \textbf{0.802} & 0.092 & 0.938 & 0.269 & 0.749 & 0.253 & 0.788 & 0.283 & 0.633 & 0.335 & 0.596 & 0.369 & 0.459 & 0.084 & 0.460 & 0.086 & 0.635 & 0.508 & 0.936 & 0.874\\
    \hline
    \multicolumn{2}{l|}{full} & 0.772 & 0.094 & 0.939 & 0.280 & 0.674 & 0.237 & 0.720 & 0.266 
    & 0.652 & 0.356 & 0.626 & 0.387 & 0.803 & 0.205 & 0.804 & 0.205 & 0.744 & 0.656 & 0.955 & 0.910\\
    \hline
    \multicolumn{2}{l|}{VAN-AD} & 0.797 & \textbf{0.108} & \textbf{0.953} & 0.414 & 0.762 & \textbf{0.293} & 0.778 & \textbf{0.297} & \textbf{0.815} & \textbf{0.622} & \textbf{0.817} & \textbf{0.661} & \textbf{0.937} & \textbf{0.559} & \textbf{0.921} & \textbf{0.538} & \textbf{0.829} & \textbf{0.687} & \textbf{0.964} & \textbf{0.924}\\
    \hline
    \end{tabular}
  }
  \label{tab:finetune}
  \end{table*}
% % Mapping and Modulation Analysis
% \begin{table*}
%   \caption{Performance comparison of different adaptive distribution strategies. The best ones are in bold.}
%   \centering
%   \small
%   \setlength{\tabcolsep}{4pt}
%   \renewcommand{\arraystretch}{1.5}
%   \resizebox{\textwidth}{!}{%
%     \begin{tabular}{ll|cc|cc|cc|cc|cc|cc}
%     \hline
%     \multicolumn{2}{l|}{Dataset} & \multicolumn{2}{c|}{CalIt2} & \multicolumn{2}{c|}{DLR} & \multicolumn{2}{c|}{GECCO} & \multicolumn{2}{c|}{PSM} & \multicolumn{2}{c|}{PUMP} & \multicolumn{2}{c}{SWAN}\\
%     \hline
%     \multicolumn{2}{l|}{Metric} & V-R & V-P & V-R & V-P & V-R & V-P & V-R & V-P & V-R & V-P & V-R & V-P\\
%     \hline
%     \multicolumn{2}{l|}{Mapping} & 0.937 & 0.288 & 0.980 & 0.486 & 0.979 & 0.432 & 0.747 & 0.540 & 0.802 & 0.223 & 0.957 & 0.913\\
%     \hline
%     \multicolumn{2}{l|}{\textbf{Modulation}} & \textbf{0.953} & \textbf{0.414} & \textbf{0.994} & \textbf{0.693} & \textbf{0.996} & \textbf{0.722} & \textbf{0.817} & \textbf{0.661} & \textbf{0.921} & \textbf{0.538} & \textbf{0.964} & \textbf{0.924}\\
%     \hline
  
%     \end{tabular}
%   }
%   \label{tab:mapping_modulation}
%   \end{table*}

\subsubsection{Interpolation Analysis}
We further study the effect of interpolation in the heatmap encoding process. As shown in Table~\ref{tab:interpolation_analysis}, bilinear and bicubic interpolation achieve comparable performance and consistently outperform nearest neighbor. This indicates that smoother interpolation is more suitable for resizing time-series heatmaps.

To verify whether interpolation distorts the original time-series information, we conduct a resize-and-restore analysis. Each window $X\in\mathbb{R}^{C\times L}$ is first resized to $224\times224$ and then restored to its original resolution as $X'$. We evaluate temporal preservation using Temporal Correlation (TC), Normalized Mean Squared Error (NMSE), and First-order Temporal Correlation (FTC). We further evaluate inter-variable structural preservation by comparing the Pearson correlation matrices of $X$ and $X'$ using Correlation Matrix Similarity (CMS), Relative Correlation Matrix Error (RCME), and Top-$k$ Neighbor Preservation Rate (TNPR).

As shown in Table~\ref{tab:interpolation_similarity}, most datasets show high TC/FTC and low NMSE, indicating that temporal dynamics are well preserved after resizing. Their inter-variable structures are also largely maintained, as suggested by high CMS and TNPR values. CICIDS is the main exception, where lower CMS/TNPR and higher RCME imply that resizing may affect its variable correlation structure, probably due to its larger number of variables. Overall, these results show that heatmap encoding with smooth interpolation preserves both temporal dynamics and inter-variable relationships in most cases.

% Density Estimation Analysis
\begin{table}
  \caption{Comparison of different anomaly scoring and density estimation modules. 
  MAE+ADMM denotes inference without NFM. 
  KDE, GMM, Deep SVDD, and MADE are alternative density estimation or anomaly scoring modules, while MTGFLOW, GANF, and NFM are normalizing-flow-based methods. 
  The best results are highlighted in bold.}
  \centering
  \small
  \setlength{\tabcolsep}{4pt}
  \renewcommand{\arraystretch}{1.5}
  \resizebox{\columnwidth}{!}{%
    \begin{tabular}{ll|cc|cc|cc|cc}
    \hline
    \multicolumn{2}{l|}{Dataset} & \multicolumn{2}{c|}{CalIt2} & \multicolumn{2}{c|}{PSM} & \multicolumn{2}{c|}{PUMP} & \multicolumn{2}{c}{SWAN}\\
    \hline
    \multicolumn{2}{l|}{Method} & V-R & V-P & V-R & V-P & V-R & V-P & V-R & V-P\\
    \hline
    \multicolumn{2}{l|}{MAE+ADMM} & 0.947 & 0.412 & 0.745 & 0.585 & 0.479 & 0.142 & 0.949 & 0.918\\
    \hline
    \multicolumn{2}{l|}{+KDE} & 0.952 & 0.413 & 0.769 & 0.609 & 0.792 & 0.218 & \textbf{0.964} & \textbf{0.934}\\
    \hline
    \multicolumn{2}{l|}{+GMM} & \textbf{0.953} & 0.413 & 0.757 & 0.602 & 0.798 & 0.218 & 0.951 & 0.910\\
    \hline
    \multicolumn{2}{l|}{+Deep SVDD} & 0.948 & 0.412 & 0.746 & 0.586 & 0.479 & 0.142 & 0.949 & 0.918\\
    \hline
    \multicolumn{2}{l|}{+MADE} & 0.952 & 0.411 & 0.811 & 0.652 & 0.819 & 0.259 & 0.952 & 0.906\\
    \hline
    \multicolumn{2}{l|}{+MTGFLOW} & 0.947 & 0.410 & 0.748 & 0.585 & 0.725 & 0.172 & 0.918 & 0.718\\
    \hline
    \multicolumn{2}{l|}{+GANF} & 0.950 & 0.412 & 0.744 & 0.579 & 0.803 & 0.214 & 0.918 & 0.803\\
    \hline
    \multicolumn{2}{l|}{\textbf{+NFM}} & \textbf{0.953} & \textbf{0.414} & \textbf{0.817} & \textbf{0.661} & \textbf{0.921} & \textbf{0.538} & \textbf{0.964} & 0.924\\
    \hline
  
    \end{tabular}
  }
  \label{tab:density_estimation}
  \end{table}

\subsubsection{Fine-tuning Analysis}
We further evaluate the effect of jointly fine-tuning MAE during training, whereas in the previous setting the MAE remains frozen. Five fine-tuning strategies are considered: (1) \textit{ln}, which fine-tunes the layer normalization layers; (2) \textit{bias}, which updates only the bias terms; (3) \textit{attn}, which fine-tunes only the self-attention related parameters; (4) \textit{mlp}, which fine-tunes only the feed-forward network components in the Transformer blocks; and (5) \textit{full}, which fine-tunes all MAE parameters. The learning rate for MAE fine-tuning is set to 0.005 by default, and each dataset is fine-tuned for 5 epochs. The results are reported in Table \ref{tab:finetune}.

We observe that fine-tuning has limited impact on datasets with relatively few variables, such as CalIt2 and MSL. In contrast, on high dimensional datasets, fine-tuning consistently degrades performance. One possible explanation is that datasets with more variables exhibit more complex inter-variable dependencies, which makes MAE more likely to overfit dataset patterns during fine-tuning. As a result, the transferable visual priors learned during pre-training are weakened, and the separability between normal and anomalous patterns is reduced.

\begin{table}
\caption{Base density experiments for VAN-AD. Fix denotes the use of a fixed base density, i.e., $p_Z(z_t)=\mathcal{N}(z_t \mid 0, I)$, whereas Rand denotes the use of a randomly initialized base density, corresponding to our default setting $p_Z(z_t)=\mathcal{N}(z_t \mid u, I)$. The best ones are in bold.
}
\centering
\small
\setlength{\tabcolsep}{4pt}
\renewcommand{\arraystretch}{1.5}
\resizebox{\columnwidth}{!}{%
  \begin{tabular}{ll|cc|cc|cc|cc}
  \hline
  \multicolumn{2}{l|}{Dataset} & \multicolumn{2}{c|}{CalIt2} & \multicolumn{2}{c|}{PSM} & \multicolumn{2}{c|}{PUMP} & \multicolumn{2}{c}{SWAN}\\
  \hline
  \multicolumn{2}{l|}{Metric} & V-R & V-P & V-R & V-P & V-R & V-P & V-R & V-P\\
  \hline
  \multicolumn{2}{l|}{Fix} & 0.952 & 0.411 & 0.800 & 0.643 & 0.794 & 0.248 & 0.961 & 0.916\\
  \hline
  \multicolumn{2}{l|}{Rand} & \textbf{0.953} & \textbf{0.414} & \textbf{0.817} & \textbf{0.661} & \textbf{0.921} & \textbf{0.538} & \textbf{0.964} & \textbf{0.924} \\
  \hline

  \end{tabular}
}
\label{tab:base_density}
\end{table}

% Parameter Sensitivity
\begin{figure}[t]
  \centering
  \includegraphics[width=1.0\linewidth]{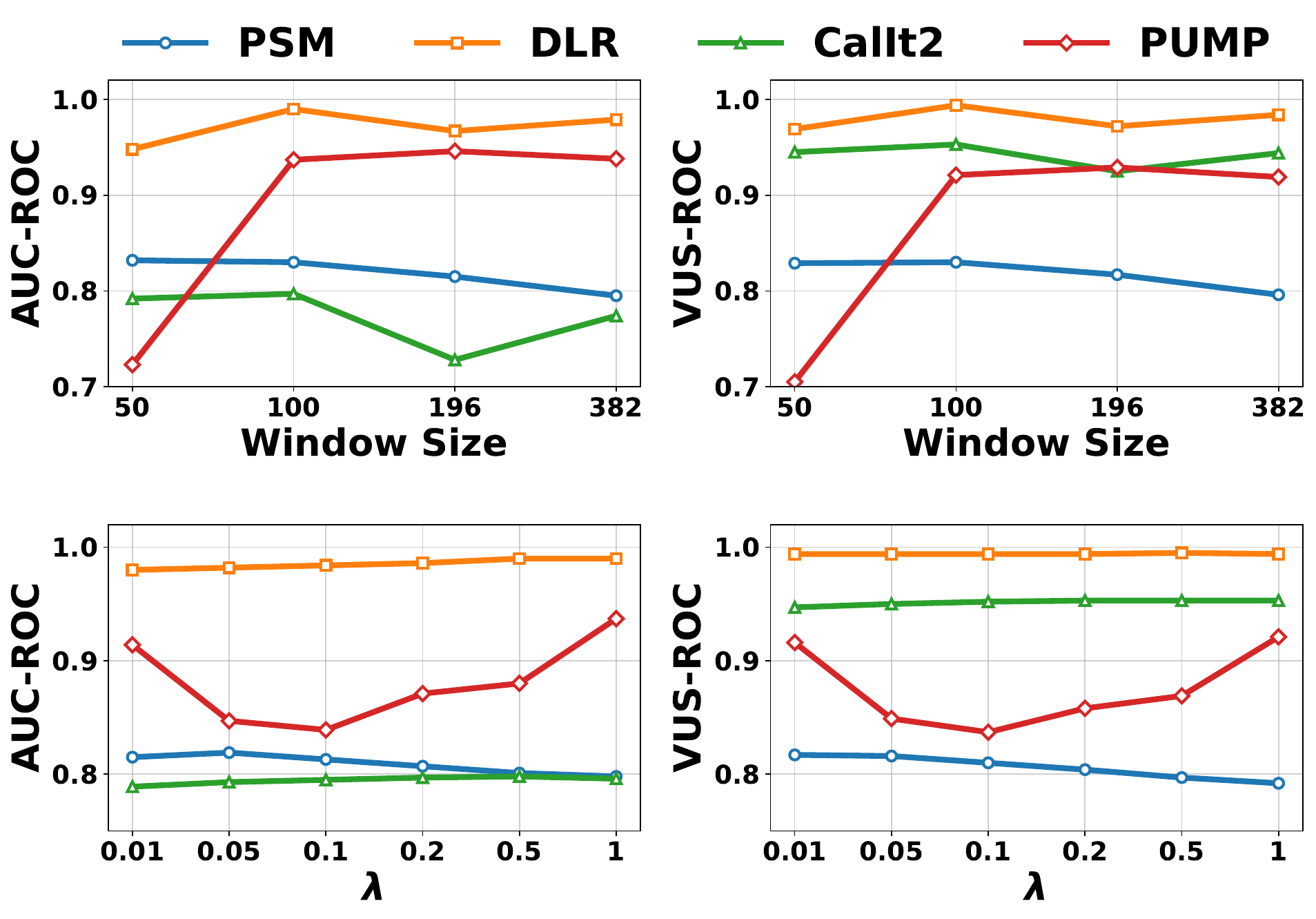}
  \caption{Parameter sensitivity studies of main hyper-parameters in VAN-AD.}
  \label{fig:parameters_sensitivity}
\end{figure}
% The Flow Analysis
\begin{table}
\caption{Comparison of different flow architectures. Linear denotes that the linear layer is implemented as a fully connected layer, whereas Masked denotes the architecture adopted in our method. The best ones are in bold.
}
\centering
\small
\setlength{\tabcolsep}{4pt}
\renewcommand{\arraystretch}{1.5}
\resizebox{\columnwidth}{!}{%
  \begin{tabular}{ll|cc|cc|cc|cc}
  \hline
  \multicolumn{2}{l|}{Dataset} & \multicolumn{2}{c|}{CalIt2} & \multicolumn{2}{c|}{PSM} & \multicolumn{2}{c|}{PUMP} & \multicolumn{2}{c}{SWAN}\\
  \hline
  \multicolumn{2}{l|}{Metric} & V-R & V-P & V-R & V-P & V-R & V-P & V-R & V-P\\
  \hline
  \multicolumn{2}{l|}{Linear} & \textbf{0.953} & 0.412 & 0.742 & 0.516 & 0.781 & 0.208 & 0.950 & 0.890\\
  \hline
  \multicolumn{2}{l|}{Masked} & \textbf{0.953} & \textbf{0.414} & \textbf{0.817} & \textbf{0.661} & \textbf{0.921} & \textbf{0.538} & \textbf{0.964} & \textbf{0.924} \\
  \hline

  \end{tabular}
}
\label{tab:flow_masked_linear}
\end{table}

% ADMM Visualization
\begin{figure*}[t]
  \centering
  \includegraphics[width=1.0\linewidth]{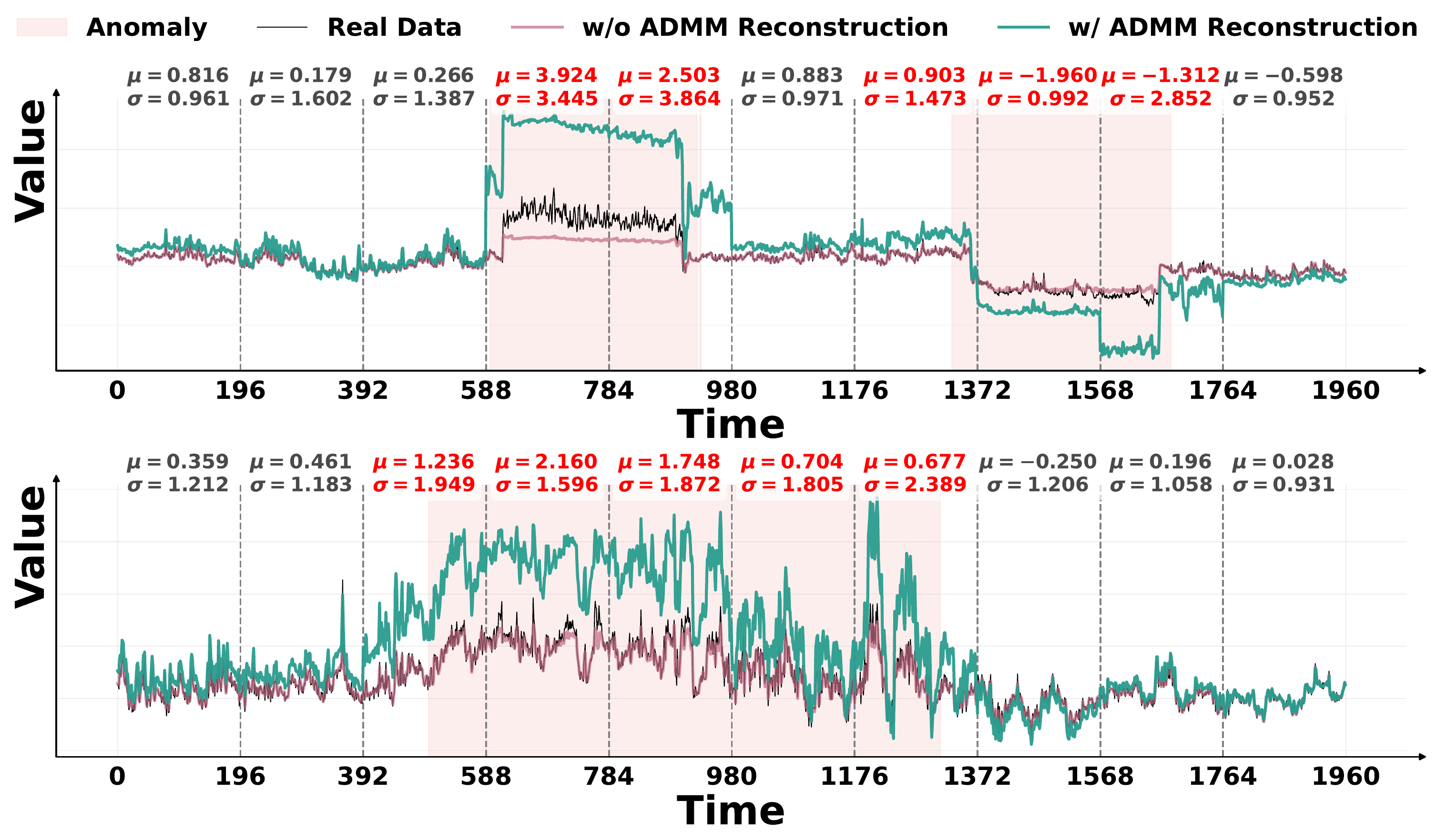}
  \caption{Visualization of the effect of ADMM on the reconstruction performance of MAE in the PSM dataset. The first and second panels show time steps 6468--8428 and 47628--49588 of the 13th variable in PSM, respectively. For each window, $\mu$ and $\sigma$ denote the local mean and standard deviation.}
  \label{fig:admm_combine}
\end{figure*}

\subsection{Further Analysis of Normalizing Flow Module}
\subsubsection{Density Modeling Comparison}
To further verify the effectiveness of NFM, we replace it with several alternative anomaly scoring and density estimation modules, including KDE~\cite{parzen1962estimation}, GMM~\cite{dempster1977maximum}, Deep SVDD~\cite{ruff2018deep}, and MADE~\cite{germain2015made}. We also compare NFM with other normalizing-flow-based methods, including GANF~\cite{dai2022graph} and MTGFLOW~\cite{zhou2024label}. The results are summarized in Table~\ref{tab:density_estimation}.

Compared with directly using MAE+ADMM, introducing density estimation consistently improves the detection performance, indicating that modeling the global distribution of normal patterns is beneficial for TSAD. However, different density estimation methods exhibit substantially different performance. Classical methods such as KDE and GMM provide only limited improvements, while Deep SVDD brings almost no gain. MADE achieves better results but still consistently underperforms NFM.

Among the flow-based methods, MTGFLOW and GANF improve the performance on some datasets but remain inferior to NFM overall. NFM achieves the best average performance across all datasets, especially on PSM and PUMP. These results suggest that the improvement is not merely due to introducing an additional trainable detector, but mainly comes from the effective density modeling capability of the proposed normalizing flow module.

\subsubsection{Base Density Analysis}
In this paper, we define the base density as a Gaussian distribution with mean vector $u \in \mathbb{R}^{C}$ and identity covariance, i.e., $p_Z(z_t)=\mathcal{N}(z_t \mid u, I)$, where $u$ is randomly initialized once at model initialization and then kept fixed. To examine the effect of this design, we replace the randomly initialized Gaussian prior with a fixed standard Gaussian prior, i.e., $p_Z(z_t)=\mathcal{N}(z_t \mid 0, I)$. As shown in Table \ref{tab:base_density}, the two settings perform similarly on datasets with relatively few variables, such as CalIt2, whereas random initialization consistently performs better on high dimensional datasets, with particularly notable gains on PSM and PUMP. One possible explanation is that, when the number of variables is small, the joint distribution is relatively simple, so both priors provide comparable alignment targets for the flow. By contrast, for high dimensional multivariate data, a fixed zero-centered prior imposes a stronger constraint on latent alignment, while random initialization offers a more flexible anchor for the reversible mapping, thereby facilitating optimization and better preserving inter-variable structure.

% The Few-shot analysis
\begin{table}
  \caption{Few-shot experiments for VAN-AD, evaluated by V-R and V-P. Zero-shot denotes using ADMM only on top of MAE. The best ones are in bold.
  }
  \centering
  \small
  \setlength{\tabcolsep}{4pt}
  \renewcommand{\arraystretch}{1.5}
  \resizebox{\columnwidth}{!}{%
    \begin{tabular}{ll|cc|cc|cc|cc}
    \hline
    \multicolumn{2}{l|}{Dataset} & \multicolumn{2}{c|}{PSM} & \multicolumn{2}{c|}{MSL} & \multicolumn{2}{c|}{SWAN} & \multicolumn{2}{c}{PUMP}\\
    \hline
    \multicolumn{2}{l|}{Metric} & V-R & V-P & V-R & V-P & V-R & V-P & V-R & V-P\\
    \hline
    \multicolumn{2}{l|}{Zero-shot} & 0.745 & 0.585 & 0.773 & 0.288 & 0.949 & 0.918 & 0.479 & 0.142\\
    \hline
    \multicolumn{2}{l|}{10\%-shot} & 0.763 & 0.605 & 0.773 & 0.288 & 0.957 & 0.915 & 0.774 & 0.214\\
    \hline
    \multicolumn{2}{l|}{20\%-shot} & 0.766 & 0.602 & 0.776 & 0.289 & 0.953 & 0.908 & 0.686 & 0.140\\
    \hline
    \multicolumn{2}{l|}{30\%-shot} & 0.789 & 0.634 & 0.776 & 0.289 & 0.953 & 0.909 & 0.725 & 0.167\\
    \hline
    \multicolumn{2}{l|}{\textbf{Full-shot}} & \textbf{0.817} & \textbf{0.661} & \textbf{0.778} & \textbf{0.297} & \textbf{0.964} & \textbf{0.924} & \textbf{0.921} & \textbf{0.538}\\
    \hline
  
    \end{tabular}
  }
  \label{tab:few_shot}
  \end{table}

% NF Visualization
\begin{figure*}[t]
  \centering
  \includegraphics[width=1.0\linewidth]{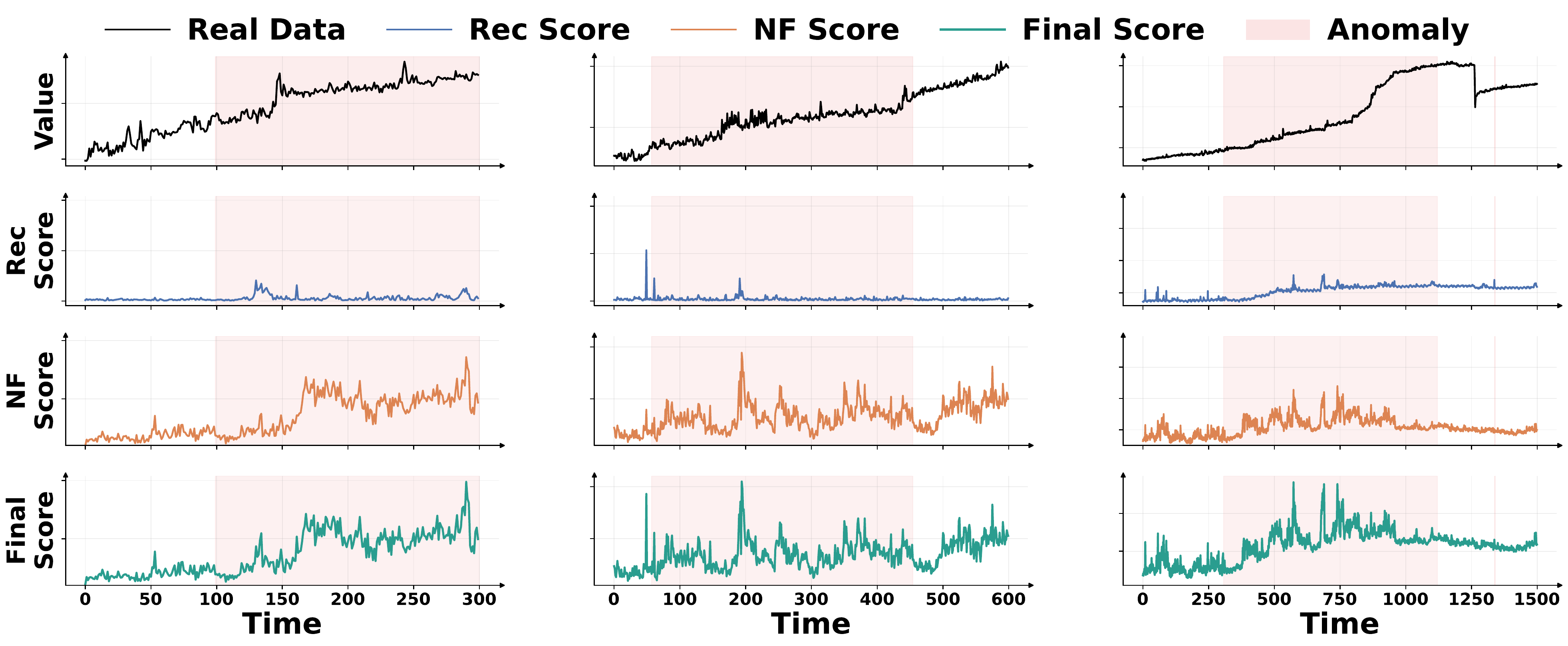}
  \caption{Visualization of the reconstruction score (Rec Score) and the anomaly score computed by normalizing flow (NF Score) for the 2nd variable in the PSM dataset. The three real data sequences correspond to time steps 7700--8000, 45300--47700, and 53000--54500, respectively.}
  \label{fig:nfm_combine}
\end{figure*}

\subsubsection{The Flow $f$ Analysis}
We also evaluate the effect of the flow architecture by replacing the Masked Linear layers with fully connected linear layers. As shown in Table \ref{tab:flow_masked_linear}, this replacement has little effect on datasets with relatively few variables, such as CalIt2, but leads to more obvious performance degradation on datasets with many variables. This suggests that the Masked Linear design is more suitable for multivariate density modeling, because its autoregressive structure with mask reversal allows inter-variable dependencies to be captured in a more structured way than fully connected layers.

\subsubsection{Few-shot Analysis}
As shown in Table \ref{tab:few_shot}, we conduct few-shot experiments by using 10\%, 20\%, and 30\% of the training data. Few-shot training generally improves performance over direct inference, with particularly clear gains on PSM and PUMP. However, the improvement is not strictly monotonic across all datasets, suggesting that few-shot adaptation can be sensitive to the amount and representativeness of the available training samples. Overall, the full-shot setting still provides the most stable and best performance.

\subsection{Parameter Sensitivity Analysis}
We also investigate the parameter sensitivity of VAN-AD. Fig. \ref{fig:parameters_sensitivity} summarizes the effects of the input window size and the balancing coefficient $\lambda$, which controls the contribution of the local and global anomaly scores. Overall, VAN-AD remains relatively stable under different window sizes and is generally robust to $\lambda$. In our experiments, the window size is typically set to 100 or 196, and good performance can be achieved when $\lambda$ is set within the range of 0.01 to 0.1.

\subsection{Visualization}
\subsubsection{The Effect of ADMM}
\label{sec:visualizaton of admm}
Fig.~\ref{fig:admm_combine} visualizes the reconstruction results of MAE with and without ADMM. 
Compared with normal windows, anomalous windows show more significant changes in local statistics, particularly enlarged standard deviations. 
ADMM uses these local statistics to recalibrate the MAE reconstruction, thereby amplifying reconstruction errors in anomalous segments and making them easier to distinguish. 
This is consistent with the theoretical analysis in Section~\ref{section:admm}, where a larger local standard deviation $\sigma_c$ strengthens reconstruction errors after ADMM.

\subsubsection{The Effect of NFM}
We further visualize the working mechanism of VAN-AD. Fig. \ref{fig:nfm_combine} presents the anomaly detection results of VAN-AD on the PSM dataset. The reconstruction scores are shown in the second row, the normalizing flow scores are shown in the third row, and the final anomaly scores are shown in the fourth row. It can be observed that when an anomalous segment is particularly long, reconstruction within the current window alone may produce low reconstruction scores, making it difficult to identify anomalies based only on reconstruction error. In contrast, NFM provides an additional anomaly signal by estimating the probability density of the current window, thereby compensating for the limited local perception of MAE.

\section{Conclusion}
In this paper, we investigate the transferability of pre-trained vision models to time series anomaly detection, offering a new perspective beyond conventional Text-based and TS-based methods. By exploiting the structural similarity between images and time series, we propose VAN-AD, a novel framework that integrates pre-trained visual masked autoencoder and normalizing flow for TSAD. To adapt MAE to this task, we design ADMM to alleviate the over-generalization issue in reconstruction, and develop NFM to model the global probability density of normal patterns, thereby addressing the limited local perception of MAE. Extensive experiments on multiple benchmark datasets demonstrate that VAN-AD consistently outperforms state-of-the-art methods across different evaluation metrics. We hope that this work can provide useful insights into cross-modal transfer for time series anomaly detection. In future work, we plan to: (1) explore anomaly generation techniques to synthesize realistic anomalous patterns, thereby alleviating class imbalance and improving anomaly detection under data-scarce settings; (2) incorporate LLM-based text modeling into the current framework to enable multi-modal anomaly detection~\cite{hu2026mindts,he2026harnessing,chowdhury2026t3time}; and (3) investigate more challenging settings, such as time series anomaly prediction, where anomalous events need to be identified before they occur~\cite{chen2025red,park2025will}.

\bibliography{ref.bib}

@article{zamanzadeh2024deep,
  title={Deep learning for time series anomaly detection: A survey},
  author={Zamanzadeh Darban, Zahra and Webb, Geoffrey I and Pan, Shirui and Aggarwal, Charu and Salehi, Mahsa},
  journal={ACM Computing Surveys},
  volume={57},
  number={1},
  pages={1--42},
  year={2024},
  publisher={ACM New York, NY}
}

@inproceedings{wucatch,
  title={CATCH: Channel-Aware Multivariate Time Series Anomaly Detection via Frequency Patching},
  author={Wu, Xingjian and Qiu, Xiangfei and Li, Zhengyu and Wang, Yihang and Hu, Jilin and Guo, Chenjuan and Xiong, Hui and Yang, Bin},
  booktitle={The Thirteenth International Conference on Learning Representations}
}

@inproceedings{licrossad,
  title={CrossAD: Time Series Anomaly Detection with Cross-scale Associations and Cross-window Modeling},
  author={Li, Beibu and Shentu, Qichao and Shu, Yang and Zhang, Hui and Li, Ming and Jin, Ning and Yang, Bin and Guo, Chenjuan},
  booktitle={The Thirty-ninth Annual Conference on Neural Information Processing Systems}
}

@inproceedings{yinscatterad,
  title={ScatterAD: Temporal-Topological Scattering Mechanism for Time Series Anomaly Detection},
  author={Yin, Tao and Fu, Shaochen and Zhang, Zhibin and Huang, Li and Zhang, Xiaohong and Yang, Yiyuan and Yang, Kaixiang and Yan, Meng},
  booktitle={The Thirty-ninth Annual Conference on Neural Information Processing Systems}
}

@inproceedings{shentu2025towards,
  title={TOWARDS A GENERAL TIME SERIES ANOMALY DETECTOR WITH ADAPTIVE BOTTLENECKS AND DUAL ADVERSARIAL DECODERS},
  author={Shentu, Qichao and Li, Beibu and Zhao, Kai and Shu, Yang and Rao, Zhongwen and Pan, Lujia and Yang, Bin and Guo, Chenjuan},
  booktitle={13th International Conference on Learning Representations, ICLR 2025},
  pages={18810--18833},
  year={2025},
  organization={International Conference on Learning Representations, ICLR}
}

@inproceedings{liu2024large,
  title={Large language model guided knowledge distillation for time series anomaly detection},
  author={Liu, Chen and He, Shibo and Zhou, Qihang and Li, Shizhong and Meng, Wenchao},
  booktitle={Proceedings of the Thirty-Third International Joint Conference on Artificial Intelligence},
  pages={2162--2170},
  year={2024}
}

@inproceedings{zhoucan,
  title={Can LLMs Understand Time Series Anomalies?},
  author={Zhou, Zihao and Yu, Rose},
  booktitle={The Thirteenth International Conference on Learning Representations}
}

@article{dong2024can,
  title={Can llms serve as time series anomaly detectors?},
  author={Dong, Manqing and Huang, Hao and Cao, Longbing},
  journal={arXiv preprint arXiv:2408.03475},
  year={2024}
}

@inproceedings{chen2025visionts,
  title={VisionTS: Visual Masked Autoencoders Are Free-Lunch Zero-Shot Time Series Forecasters},
  author={Chen, Mouxiang and Shen, Lefei and Li, Zhuo and Wang, Xiaoyun Joy and Sun, Jianling and Liu, Chenghao},
  booktitle={International Conference on Machine Learning},
  pages={8979--9007},
  year={2025},
  organization={PMLR}
}

@article{shen2025visionts++,
  title={VisionTS++: Cross-Modal Time Series Foundation Model with Continual Pre-trained Vision Backbones},
  author={Shen, Lefei and Chen, Mouxiang and Liu, Xu and Fu, Han and Ren, Xiaoxue and Sun, Jianling and Li, Zhuo and Liu, Chenghao},
  journal={arXiv preprint arXiv:2508.04379},
  year={2025}
}

@article{li2023time,
  title={Time series as images: Vision transformer for irregularly sampled time series},
  author={Li, Zekun and Li, Shiyang and Yan, Xifeng},
  journal={Advances in Neural Information Processing Systems},
  volume={36},
  pages={49187--49204},
  year={2023}
}

@inproceedings{he2022masked,
  title={Masked autoencoders are scalable vision learners},
  author={He, Kaiming and Chen, Xinlei and Xie, Saining and Li, Yanghao and Doll{\'a}r, Piotr and Girshick, Ross},
  booktitle={Proceedings of the IEEE/CVF conference on computer vision and pattern recognition},
  pages={16000--16009},
  year={2022}
}

@inproceedings{shen2025learn,
  title={Learn hybrid prototypes for multivariate time series anomaly detection},
  author={Shen, Ke-Yuan},
  booktitle={The Thirteenth International Conference on Learning Representations}
}

@article{song2023memto,
  title={Memto: Memory-guided transformer for multivariate time series anomaly detection},
  author={Song, Junho and Kim, Keonwoo and Oh, Jeonglyul and Cho, Sungzoon},
  journal={Advances in Neural Information Processing Systems},
  volume={36},
  pages={57947--57963},
  year={2023}
}

@article{haq2025transnas,
  title={TransNAS-TSAD: harnessing transformers for multi-objective neural architecture search in time series anomaly detection},
  author={Haq, Ijaz Ul and Lee, Byung Suk and Rizzo, Donna M},
  journal={Neural Computing and Applications},
  volume={37},
  number={4},
  pages={2455--2477},
  year={2025},
  publisher={Springer}
}

@inproceedings{park2026paano,
  title={PaAno: Patch-based Representation Learning for Time-Series Anomaly Detection},
  author={Park, Jinju and Kang, Seokho},
  booktitle={Proceedings of International Conference on Learning Representations},
  year={2026}
}

@article{vaswani2017attention,
  title={Attention is all you need},
  author={Vaswani, Ashish and Shazeer, Noam and Parmar, Niki and Uszkoreit, Jakob and Jones, Llion and Gomez, Aidan N and Kaiser, {\L}ukasz and Polosukhin, Illia},
  journal={Advances in neural information processing systems},
  volume={30},
  year={2017}
}

@inproceedings{breunig2000lof,
  title={LOF: identifying density-based local outliers},
  author={Breunig, Markus M and Kriegel, Hans-Peter and Ng, Raymond T and Sander, J{\"o}rg},
  booktitle={Proceedings of the 2000 ACM SIGMOD international conference on Management of data},
  pages={93--104},
  year={2000}
}

@article{he2003discovering,
  title={Discovering cluster-based local outliers},
  author={He, Zengyou and Xu, Xiaofei and Deng, Shengchun},
  journal={Pattern recognition letters},
  volume={24},
  number={9-10},
  pages={1641--1650},
  year={2003},
  publisher={Elsevier}
}

@article{shyu2003novel,
  title={A novel anomaly detection scheme based on principal component classifier},
  author={Shyu, Mei-Ling and Chen, Shu-Ching and Sarinnapakorn, Kanoksri and Chang, LiWu},
  year={2003}
}

@inproceedings{ramaswamy2000efficient,
  title={Efficient algorithms for mining outliers from large data sets},
  author={Ramaswamy, Sridhar and Rastogi, Rajeev and Shim, Kyuseok},
  booktitle={Proceedings of the 2000 ACM SIGMOD international conference on Management of data},
  pages={427--438},
  year={2000}
}

@inproceedings{deng2021graph,
  title={Graph neural network-based anomaly detection in multivariate time series},
  author={Deng, Ailin and Hooi, Bryan},
  booktitle={Proceedings of the AAAI conference on artificial intelligence},
  volume={35},
  number={5},
  pages={4027--4035},
  year={2021}
}

@inproceedings{xie2025multivariate,
  title={Multivariate time series anomaly detection by capturing coarse-grained intra-and inter-variate dependencies},
  author={Xie, Yongzheng and Zhang, Hongyu and Babar, Muhammad Ali},
  booktitle={Proceedings of the ACM on Web Conference 2025},
  pages={697--705},
  year={2025}
}

@inproceedings{yang2023dcdetector,
  title={Dcdetector: Dual attention contrastive representation learning for time series anomaly detection},
  author={Yang, Yiyuan and Zhang, Chaoli and Zhou, Tian and Wen, Qingsong and Sun, Liang},
  booktitle={Proceedings of the 29th ACM SIGKDD conference on knowledge discovery and data mining},
  pages={3033--3045},
  year={2023}
}

@article{kim2025causality,
  title={Causality-aware contrastive learning for robust multivariate time-series anomaly detection},
  author={Kim, HyunGi and Mok, Jisoo and Lee, Dongjun and Lew, Jaihyun and Kim, Sungjae and Yoon, Sungroh},
  journal={arXiv preprint arXiv:2506.03964},
  year={2025}
}

@inproceedings{xiaoming2025time,
  title={Time-MoE: Billion-Scale Time Series Foundation Models with Mixture of Experts},
  author={Xiaoming, Shi and Shiyu, Wang and Yuqi, Nie and Dianqi, Li and Zhou, Ye and Qingsong, Wen and Jin, Ming},
  booktitle={ICLR 2025: The Thirteenth International Conference on Learning Representations},
  year={2025},
  organization={International Conference on Learning Representations}
}

@article{ansari2024chronos,
  title={Chronos: Learning the Language of Time Series},
  author={Ansari, Abdul Fatir and Stella, Lorenzo and Turkmen, Caner and Zhang, Xiyuan and Mercado, Pedro and Shen, Huibin and Shchur, Oleksandr and Rangapuram, Syama Sundar and Arango, Sebastian Pineda and Kapoor, Shubham and others},
  journal={Transactions on Machine Learning Research},
  volume={2024},
  year={2024},
  publisher={Transactions on Machine Learning Research}
}

@inproceedings{liu2024timer,
  title={Timer: generative pre-trained transformers are large time series models},
  author={Liu, Yong and Zhang, Haoran and Li, Chenyu and Huang, Xiangdong and Wang, Jianmin and Long, Mingsheng},
  booktitle={Proceedings of the 41st International Conference on Machine Learning},
  pages={32369--32399},
  year={2024}
}

@article{zhou2023one,
  title={One fits all: Power general time series analysis by pretrained lm},
  author={Zhou, Tian and Niu, Peisong and Sun, Liang and Jin, Rong and others},
  journal={Advances in neural information processing systems},
  volume={36},
  pages={43322--43355},
  year={2023}
}

@inproceedings{zhang2024large,
  title={Large language models for spatial trajectory patterns mining},
  author={Zhang, Zheng and Amiri, Hossein and Liu, Zhenke and Zhao, Liang and Z{\"u}fle, Andreas},
  booktitle={Proceedings of the 1st ACM SIGSPATIAL International Workshop on Geospatial Anomaly Detection},
  pages={52--55},
  year={2024}
}

@article{gong2021ast,
  title={Ast: Audio spectrogram transformer},
  author={Gong, Yuan and Chung, Yu-An and Glass, James},
  journal={arXiv preprint arXiv:2104.01778},
  year={2021}
}

@inproceedings{touvron2021training,
  title={Training data-efficient image transformers \& distillation through attention},
  author={Touvron, Hugo and Cord, Matthieu and Douze, Matthijs and Massa, Francisco and Sablayrolles, Alexandre and J{\'e}gou, Herv{\'e}},
  booktitle={International conference on machine learning},
  pages={10347--10357},
  year={2021},
  organization={PMLR}
}

@inproceedings{ni2025harnessing,
  title={Harnessing Vision Models for Time Series Analysis: A Survey},
  author={Ni, Jingchao and Zhao, Ziming and Shen, Cheng Ao and Tong, Hanghang and Song, Dongjin and Cheng, Wei and Luo, Dongsheng and Chen, Haifeng},
  booktitle={34th Internationa Joint Conference on Artificial Intelligence, IJCAI 2025},
  pages={10612--10620},
  year={2025},
  organization={International Joint Conferences on Artificial Intelligence}
}

@article{zhao2025images,
  title={From Images to Signals: Are Large Vision Models Useful for Time Series Analysis?},
  author={Zhao, Ziming and Shen, ChengAo and Tong, Hanghang and Song, Dongjin and Deng, Zhigang and Wen, Qingsong and Ni, Jingchao},
  journal={arXiv preprint arXiv:2505.24030},
  year={2025}
}

@inproceedings{deng2009imagenet,
  title={Imagenet: A large-scale hierarchical image database},
  author={Deng, Jia and Dong, Wei and Socher, Richard and Li, Li-Jia and Li, Kai and Fei-Fei, Li},
  booktitle={2009 IEEE conference on computer vision and pattern recognition},
  pages={248--255},
  year={2009},
  organization={Ieee}
}

@inproceedings{dai2022graph,
  title={Graph-augmented normalizing flows for anomaly detection of multiple time series},
  author={Dai, Enyan and Chen, Jie},
  booktitle={International Conference on Learning Representations},
  year={2022}
}

@article{zhou2024label,
  title={Label-free multivariate time series anomaly detection},
  author={Zhou, Qihang and He, Shibo and Liu, Haoyu and Chen, Jiming and Meng, Wenchao},
  journal={IEEE Transactions on Knowledge and Data Engineering},
  volume={36},
  number={7},
  pages={3166--3179},
  year={2024},
  publisher={IEEE}
}

@article{papamakarios2017masked,
  title={Masked autoregressive flow for density estimation},
  author={Papamakarios, George and Pavlakou, Theo and Murray, Iain},
  journal={Advances in neural information processing systems},
  volume={30},
  year={2017}
}

@inproceedings{liu2025calf,
  title={Calf: Aligning llms for time series forecasting via cross-modal fine-tuning},
  author={Liu, Peiyuan and Guo, Hang and Dai, Tao and Li, Naiqi and Bao, Jigang and Ren, Xudong and Jiang, Yong and Xia, Shu-Tao},
  booktitle={Proceedings of the AAAI Conference on Artificial Intelligence},
  volume={39},
  number={18},
  pages={18915--18923},
  year={2025}
}

@inproceedings{liuitransformer,
  title={iTransformer: Inverted Transformers Are Effective for Time Series Forecasting},
  author={Liu, Yong and Hu, Tengge and Zhang, Haoran and Wu, Haixu and Wang, Shiyu and Ma, Lintao and Long, Mingsheng},
  booktitle={The Twelfth International Conference on Learning Representations}
}

@inproceedings{wangd3r,
  author= {Chengsen Wang and Zirui Zhuang and Qi Qi and Jingyu Wang and Xingyu Wang and Haifeng Sun and Jianxin Liao},
  title     = {Drift doesn't matter: Dynamic decomposition with diffusion reconstruction for unstable multivariate time series anomaly detection},
  booktitle = {Thirty-seventh Conference on Neural Information Processing Systems},
  year      = {2023},
}

@inproceedings{huet2022local,
  title={Local evaluation of time series anomaly detection algorithms},
  author={Huet, Alexis and Navarro, Jose Manuel and Rossi, Dario},
  booktitle={Proceedings of the 28th ACM SIGKDD Conference on Knowledge Discovery and Data Mining},
  pages={635--645},
  year={2022}
}

@article{liu2024elephant,
  title={The elephant in the room: Towards a reliable time-series anomaly detection benchmark},
  author={Liu, Qinghua and Paparrizos, John},
  journal={Advances in Neural Information Processing Systems},
  volume={37},
  pages={108231--108261},
  year={2024}
}

@inproceedings{qiu2025tab,
title      = {{TAB}: Unified Benchmarking of Time Series Anomaly Detection Methods},
author     = {Xiangfei Qiu and Zhe Li and Wanghui Qiu and Shiyan Hu and Lekui Zhou and Xingjian Wu and Zhengyu Li and Chenjuan Guo and Aoying Zhou and Zhenli Sheng and Jilin Hu and Christian S. Jensen and Bin Yang},
booktitle  = {Proc. {VLDB} Endow.},
year       = {2025}
}

@inproceedings{hu2026mindts,
  title={Towards Multimodal Time Series Anomaly Detection with Semantic Alignment and Condensed Interaction},
  author={Hu, Shiyan and Jin, Jianxin and Shu, Yang and Chen, Peng and Yang, Bin and Guo, Chenjuan},
  journal={ICLR},
  year={2026}
}

@inproceedings{he2026harnessing,
  title={Harnessing vision-language models for time series anomaly detection},
  author={He, Zelin and Alnegheimish, Sarah and Reimherr, Matthew},
  booktitle={Proceedings of the AAAI Conference on Artificial Intelligence},
  volume={40},
  number={26},
  pages={21690--21698},
  year={2026}
}

@inproceedings{chowdhury2026t3time,
  title={T3time: Tri-modal time series forecasting via adaptive multi-head alignment and residual fusion},
  author={Chowdhury, Abdul Monaf and Akter, Rabeya and Arib, Safaeid Hossain},
  booktitle={Proceedings of the AAAI Conference on Artificial Intelligence},
  volume={40},
  number={25},
  pages={20597--20605},
  year={2026}
}

@inproceedings{fu2022mad,
  title={Mad: Self-supervised masked anomaly detection task for multivariate time series},
  author={Fu, Yiwei and Xue, Feng},
  booktitle={2022 International Joint Conference on Neural Networks (IJCNN)},
  pages={1--8},
  year={2022},
  organization={IEEE}
}

@inproceedings{fang2024temporal,
  title={Temporal-frequency masked autoencoders for time series anomaly detection},
  author={Fang, Yuchen and Xie, Jiandong and Zhao, Yan and Chen, Lu and Gao, Yunjun and Zheng, Kai},
  booktitle={2024 IEEE 40th international conference on data engineering (ICDE)},
  pages={1228--1241},
  year={2024},
  organization={IEEE}
}

@article{tang2024mlp,
  title={Mlp-mixer based masked autoencoders are effective, explainable and robust for time series anomaly detection},
  author={Tang, Qideng and Dai, Chaofan and Wu, Yahui and Zhou, Haohao},
  journal={Proceedings of the VLDB Endowment},
  volume={18},
  number={3},
  pages={798--811},
  year={2024},
  publisher={VLDB Endowment}
}

@article{liu2025time,
  title={Time series masked autoencoder for process anomaly detection of reduced information redundancy},
  author={Liu, Yi and Hu, Junhao and Jia, Mingwei and Zhou, Le},
  journal={Results in Engineering},
  pages={108590},
  year={2025},
  publisher={Elsevier}
}

@article{parzen1962estimation,
  title={On estimation of a probability density function and mode},
  author={Parzen, Emanuel},
  journal={The annals of mathematical statistics},
  volume={33},
  number={3},
  pages={1065--1076},
  year={1962},
  publisher={JSTOR}
}

@article{dempster1977maximum,
  title={Maximum likelihood from incomplete data via the EM algorithm},
  author={Dempster, Arthur P and Laird, Nan M and Rubin, Donald B},
  journal={Journal of the royal statistical society: series B (methodological)},
  volume={39},
  number={1},
  pages={1--22},
  year={1977},
  publisher={Wiley Online Library}
}

@inproceedings{ruff2018deep,
  title={Deep one-class classification},
  author={Ruff, Lukas and Vandermeulen, Robert and Goernitz, Nico and Deecke, Lucas and Siddiqui, Shoaib Ahmed and Binder, Alexander and M{\"u}ller, Emmanuel and Kloft, Marius},
  booktitle={International conference on machine learning},
  pages={4393--4402},
  year={2018},
  organization={PMLR}
}

@inproceedings{germain2015made,
  title={Made: Masked autoencoder for distribution estimation},
  author={Germain, Mathieu and Gregor, Karol and Murray, Iain and Larochelle, Hugo},
  booktitle={International conference on machine learning},
  pages={881--889},
  year={2015},
  organization={PMLR}
}

@article{chen2025red,
  title={RED-F: Reconstruction-Elimination based Dual-stream Contrastive Forecasting for Multivariate Time Series Anomaly Prediction},
  author={Chen, PengYu and Shi, Xiaohou and Chang, Yuan and Sun, Yan and Das, Sajal K},
  journal={arXiv preprint arXiv:2511.20044},
  year={2025}
}

@article{park2025will,
  title={When Will It Fail?: Anomaly to Prompt for Forecasting Future Anomalies in Time Series},
  author={Park, Min-Yeong and Lee, Won-Jeong and Kim, Seong Tae and Park, Gyeong-Moon},
  journal={arXiv preprint arXiv:2506.23596},
  year={2025}
}
\bibliographystyle{ieeetr}

\end{document}